\documentclass[10pt,twocolumn,letterpaper]{article}

\usepackage[pagenumbers]{cvpr} 

\usepackage{graphicx}
\usepackage{amsmath}
\usepackage{amssymb}
\usepackage{booktabs}
\usepackage{cite}
\usepackage{pifont}
\usepackage{threeparttable}
\usepackage{multirow}
\usepackage{tikz}
\usepackage{colortbl}
\usepackage{rotate}
\usepackage{makecell}
\usepackage{amssymb}

\usepackage[pagebackref,breaklinks,colorlinks]{hyperref}

\usepackage[capitalize]{cleveref}
\crefname{section}{Sec.}{Secs.}
\Crefname{section}{Section}{Sections}
\Crefname{table}{Table}{Tables}
\crefname{table}{Tab.}{Tabs.}

\def\cvprPaperID{4476} 
\def\confName{CVPR}
\def\confYear{2024}

\usetikzlibrary{shadows}
\usetikzlibrary{plotmarks}
\usetikzlibrary{shapes}

\definecolor{darkcyan}{rgb}{0,.79,1} 
\definecolor{LightCyan}{rgb}{0.88,1,1} 

\definecolor{earthyellow}{rgb}{0.92, 0.74, 0.40}

\definecolor{highblue}{rgb}{0, 0.54, 0.9}
\definecolor{purple}{rgb}{0.8, 0.62, 1} 

\newcommand{\mytriangle}[1]{\tikz{\node[draw=#1, color=black, fill=#1,minimum
width=0.25cm,minimum height=0.25cm,inner sep=0pt, drop shadow, very thick] at (0,0) {};}}

\newcommand{\PurpleRect}{\protect\mytriangle{purple}}
\newcommand{\BlueRect}{\protect\mytriangle{highblue}}
\newcommand{\YellowRect}{\protect\mytriangle{earthyellow}}

\def\datasetname{MM-AU}
\def\modelname{AdVersa-SD}

\begin{document}

\title{Abductive Ego-View Accident Video Understanding for Safe Driving Perception}

\author{Jianwu Fang$^{1}$, Lei-lei Li$^{2}$, Junfei Zhou$^{2}$, Junbin Xiao$^{3}$, Hongkai Yu$^{4}$, Chen Lv$^{5}$, \\Jianru Xue$^{1}$, and Tat-Seng Chua$^{3}$ \\
\small{$^1$Xi'an Jiaotong University\quad$^2$Chang'an University \quad$^3$National University of Singapore }\\\small{$^4$Cleveland State University \quad$^5$Nanyang Technological University}\\
{\tt\footnotesize 1.\{fangjianwu, jrxue\}@mail.xjtu.edu.cn} \quad{\tt\footnotesize 2.\{jeffreychou777,670160532lileilei\}@gmail.com} \\{\tt\footnotesize 3.\{junbin, chuats\}@comp.nus.edu.sg} \quad{\tt\footnotesize 4.h.yu19@csuohio.edu}\quad{\tt\footnotesize 5.lyuchen@ntu.edu.sg}}
\maketitle
\begin{abstract}
We present \textbf{\datasetname}, a novel dataset for \underline{M}ulti-\underline{M}odal \underline{A}ccident video \underline{U}nderstanding. \datasetname~contains 11,727 in-the-wild ego-view accident videos, each with temporally aligned text descriptions. We annotate over 2.23 million object boxes and 58,650 pairs of video-based accident reasons, covering 58 accident categories. \datasetname~supports various accident understanding tasks, particularly multimodal video diffusion to understand accident cause-effect chains for safe driving. 
With \datasetname, we present an \underline{A}b\underline{d}uctive accident \underline{V}ideo und\underline{ers}t\underline{a}nding framework for \underline{S}afe \underline{D}riving perception (\textbf{\modelname}).
\modelname~performs video diffusion via an Object-Centric Video Diffusion (OAVD) method which is driven by an abductive CLIP model. This model involves a contrastive interaction loss to learn the pair co-occurrence of normal, near-accident, accident frames with the corresponding text descriptions, such as accident reasons, prevention advice, and accident categories. OAVD enforces the causal region learning while fixing the content of the original frame background in video generation, to find the dominant cause-effect chain for certain accidents. Extensive experiments verify the abductive ability of \modelname~and the superiority of OAVD against the state-of-the-art diffusion models. Additionally, we provide careful benchmark evaluations for object detection and accident reason answering since \modelname~relies on precise object and accident reason information. The dataset and code are released at \url{www.lotvsmmau.net}.
 \end{abstract}

\begin{figure*}[t]
  \centering
\includegraphics[width=0.95\linewidth]{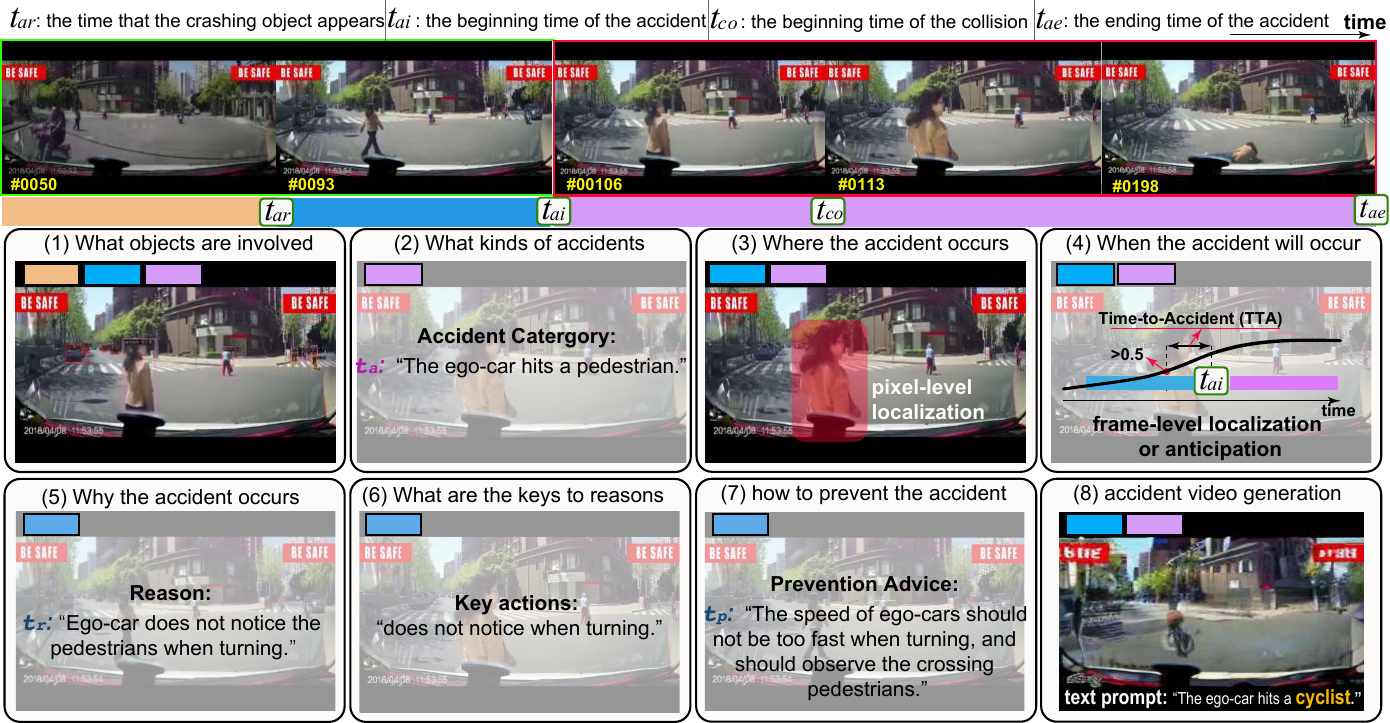}
   \caption{The ego-view multimodality accident video understanding tasks that \textbf{\datasetname} can support, where we highlight the text descriptions for accident reason (\textcolor{highblue}{$t_r$}), prevention advice (\textcolor{highblue}{$t_p$}), and accident category (\textcolor{magenta}{$t_a$}), as well as temporal windows (\textbf{accident-free}~\YellowRect, \textbf{near-accident}~\BlueRect, and \textbf{accident} \PurpleRect~windows) for different tasks.}
   \label{fig1}
   \vspace{-1.5em}
\end{figure*}
\section{Introduction}
\label{sec:intro}
Autonomous Vehicles (AV) are around the corner for practical use \cite{nature2022}. Yet, occasionally emerging traffic accidents are among the biggest obstacles to be crossed. To make a step further, 
it is urgent to comprehensively understand the traffic accidents, such as telling what objects are involved, why an accident occurs and how to prevent it. Techniques that can answer these questions are of crucial importance for safe AV systems. 
So far, there is a lack of a large-scale dataset to develop such techniques.

Therefore, this paper constructs \datasetname, a multi-modal dataset for ego-view accident video understanding. \datasetname~contains \textbf{11,727} in-the-wild ego-view accident videos. The videos are temporally aligned with the text descriptions of accident reasons, prevention solutions, and accident categories. In total, \textbf{58.6K} pairs of video-based Accident reason Answers (ArA) are annotated for 58 accident categories.  
Moreover, to enable object-centric accident video understanding, we annotate over \textbf{2.23M} object boxes for about \textbf{463K} video frames.  As shown in Fig. \ref{fig1}, \datasetname~ can facilitate 8 tasks of traffic accident video understanding, and the models are required to infer \ding{172} what objects are involved, \ding{173} what kinds of accidents, \ding{174} where and \ding{175} when the accident will occur, \ding{176} why the accident occurs, \ding{177} what are the keys to accident reasons, \ding{178} how to prevent it, and \ding{179} multimodal accident video diffusion.

Different from previous works that concentrate on the former 4 basic tasks \cite{DBLP:conf/iccv/Bao0K21,DBLP:journals/tits/FangYQXY22,wang2023gsc,kang2022vision}, we advocate an \underline{A}b\underline{d}uctive Accident \underline{V}id\underline{e}o Under\underline{s}t\underline{a}nding for \underline{S}afe \underline{D}riving (\textbf{AdVersa-SD}) perception by considering the accident reasons into the accident video understanding. For the  \ding{176}-\ding{178} tasks, few works \cite{DBLP:conf/cvpr/XuHL21,DBLP:journals/pami/LiuLL23} formulate Video Question Answering (VQA) problem to discern the accident reasons for given videos. However, understanding the cause-effect chain of accident is more crucial to prevent collision. 
Hence, we present AdVersa-SD, which underscores a  
diffusion technique to bridge the visual content (effect) with specific text prompts (cause).

Leveraging the text-vision CLIP model \cite{radford2021learning} and the video diffusion techniques \cite{voleti2022mcvd,esser2023structure}, we propose an abductive CLIP in AdVersa-SD with a constrastive interaction loss for the accident reason involved semantic co-occurrence learning within the text and video clips, such as the pairs of (\BlueRect,\textcolor{highblue}{$t_r$}) and (\PurpleRect, \textcolor{magenta}{$t_a$}). To verify the abductive ability of abductive CLIP, we treat it as an engine to drive an Object-centric Accident Video Diffusion (OAVD) model by enforcing the learning of object locations and restricting the influence of frame background for causal region generation. OAVD takes the stable diffusion model \cite{rombach2022high} as the baseline and extends it to the high-quality accident video generation by 3D convolution and well-designed spatial-temporal attention modules, as well as an object region masked latent representation reconstruction. This formulation is useful for finding the dominant cause-effect chain for certain accidents irrelevant to the environmental background. Thus, the \textbf{contributions} are threefold.
\vspace{-0.2em}

(1) A new large-scale ego-view multimodality accident understanding dataset, \emph{i.e.}, \datasetname, is created, which will facilitate the more promising abductive understanding for safe driving perception.
(2) We present AdVersa-SD, an abductive accident video understanding framework, to learn the dominant reason-occurrence elements of accidents within text-video pairs.
(3) Within AdVersa-SD, we propose an Object-centric Accident Video Diffusion (OAVD) driven by the abductive CLIP to attempt to explicitly explore the causal-effect chain of accident occurrence, and positive results are obtained.

\section{Related Work}
\subsection{Ego-View Accident Video Understanding}

\textbf{Accident Detection:}
Accident detection in ego-view videos aims to localize the spatial regions and temporal frame windows where the accident occurs. Because of the drastic change in the shape, location, and relations of the participants, the key problems of accident detection are to extract robust appearance or motion features for the representation of video frames, spatial-temporal video volumes, or trajectories. Commonly, the frame consistency \cite{DBLP:journals/tits/SinghM19,DBLP:journals/ict-express/PawarA22,zhou2022spatio,DBLP:conf/avss/HajriF22}, location consistency \cite{le2020attention,taccari2018classification,santhosh2021vehicular,DBLP:journals/eswa/HuWCG21}, and scene context consistency (\eg, the object interactions)  \cite{roy2020detection,fang2022traffic,DBLP:journals/tits/VijayDCNK23} are modeled to find the accident window or regions. Up to now, unsupervised location or frame prediction has been a common choice for model designing. For example, DoTA \cite{DBLP:conf/iros/YaoXWCA19,yao2022dota}, the typical ego-view accident detection method, computes the Standard Deviation (STD) of predicted locations of the pre-detected objects. 

\textbf{Accident Anticipation:}
Accident anticipation aims to forecast the probability and prefers an early warning of future accidents based on the complex scene structure modeling in video frames \cite{bao2020uncertainty,karim2022dynamic,DBLP:journals/iotj/MalawadeYHMKF22}. The earliness is maintained by taking the exponential loss \cite{DBLP:conf/accv/ChanCXS16,DBLP:conf/icra/JainSKSS16,suzuki2018anticipating,wang2023gsc,DBLP:conf/iccv/Bao0K21} to penalize the positive occurrence of the accident. Different from accident detection, most accident anticipation works need to provide the accident window annotation to fulfill supervised learning. One groundbreaking work by Chan \emph{et al.} \cite{DBLP:conf/accv/ChanCXS16} models a Dynamic-Spatial-Attention Recurrent Neural Network (DSA-RNN) to correlate the temporal consistency of road participants' tracklets, which is extended by the works \cite{DBLP:journals/tits/KarimLQY22,karim2022attention} to compute the riskiness of video frames or objects. To boost the explainability of accident anticipation, Bao and Kong \cite{DBLP:conf/iccv/Bao0K21} develop a Deep Reinforced accident anticipation with Visual Explanation (DRIVE) assisted by driver attention \cite{DBLP:journals/tits/FangYQXY22} and obtain a significant improvement.

\textbf{Accident Classification:}
Because of the video data limitation of different accident categories, there is a paucity of research on ego-view video-based accident classification, and many works concentrate on the surveillance view with limited image set \cite{kumeda2019vehicle, ghosh2019accident,luoICASSP2023}. Kang \emph{et al.} \cite{kang2022vision} propose a Vision Transformer-Traffic Accident (ViT-TA) model that classifies the ego-view traffic accident scenes, highlighting key objects through attention maps to increase the objectivity and reliability of functional scenes.

The aforementioned works focus on the monocular vision modal, while the spatial or temporal causal part of the accident video is hard to learn effectively owning to the complex evolution of accidents. 
\begin{table}[!t]\footnotesize
  \centering
  \caption{Attribute comparison of ego-view accident video datasets. }
 \setlength{\tabcolsep}{0.3mm}{
\begin{tabular}{c|cccccccc}
    \toprule
Datasets &Years&\#Clips &\#Frames &Bboxes&Tracklet&TA.&TT&R/S\\
\hline
DAD \cite{DBLP:conf/accv/ChanCXS16}&2016& 1,750& 175K&  &  \checkmark & \checkmark&&R\\
A3D  \cite{DBLP:conf/iros/YaoXWCA19} &2019& 3,757& 208K& & & \checkmark&&R\\
GTACrash \cite{DBLP:conf/aaai/KimLHS19} &2019& 7,720&-&  &  & \checkmark &&S\\
VIENA$^2$ \cite{aliakbarian2019viena} &2019&15,000&2.25M&  &  & \checkmark &&S\\
CTA \cite{DBLP:conf/eccv/YouH20}&2020&1,935&-&&  & \checkmark&\checkmark&R\\
CCD \cite{bao2020uncertainty}  &2021& 1,381&75K&   & \checkmark & \checkmark&&R\\
TRA \cite{DBLP:journals/tits/LiuLCLX22} &2022& 560&-&  &  & \checkmark &&R\\
DADA-2000 \cite{DBLP:journals/tits/FangYQXY22}&2022& 2000&658k&  &  & \checkmark &&R\\
DoTA \cite{yao2022dota}  &2022& 5,586&732K& partial   &  & \checkmark&&R\\
ROL \cite{karim2022attention}  &2023& 1000&100K&   &  & \checkmark&&R\\
DeepAccident \cite{DBLP:journals/corr/abs-2304-01168} &2023&-&57k&   &  & \checkmark&&S\\
CTAD \cite{luoICASSP2023} &2023&1,100&-&   &  & \checkmark&&S\\
\textbf{\datasetname} &2023& \textbf{11,727}&\textbf{2.19M}& \checkmark & & \checkmark&\checkmark&R\\
    \toprule
  \end{tabular}}
    \begin{tablenotes}
\item \scriptsize{\textbf{Bboxes}: bounding boxes of objects, \textbf{TA.}: temporal annotation of the accident, \textbf{TT}: text descriptions, \textbf{R/S}: real or synthetic datasets.}
\end{tablenotes}
  \label{tab1}
  \end{table}
  
\textbf{Accident Reason Answering:}
Closely related to this work, You and Han \cite{DBLP:conf/eccv/YouH20} investigate the causal-effect recognition of accident scenarios, and build the class taxonomy of traffic accidents. Besides, SUTD-TrafficQA \cite{DBLP:conf/cvpr/XuHL21} formulates the reason explanation and prevention advice of accidents by the Question-Answering (QA) framework, which involves the reasoning of dynamic and complex traffic scenes. Based on this, Liu \emph{et al.} \cite{DBLP:journals/pami/LiuLL23} reason the cross-modal causal relation to fulfill the traffic accident reason answering. 
We believe QA frameworks \cite{DBLP:conf/cvpr/ZangWPL23,DBLP:journals/pami/XiaoZYLHYC23} can provide a direct understanding for telling why the accident occurs. However, there is no explicit double-check solution to verify what key elements (\eg, specific actions or objects) are dominant for subsequent accidents.

\subsection{Ego-View Accident Understanding Datasets}
The community has realized the importance of accident video understanding for safe driving perception, and some benchmarks have been released in recent years. Tab. \ref{tab1} presents the attribute comparison of available ego-view accident video datasets. DAD \cite{DBLP:conf/accv/ChanCXS16} is the pioneering dataset, where each video clip is trimmed with 10 accident frames at the end of each clip. This setting is also adopted in the CCD datasets \cite{bao2020uncertainty} with a total of 50 frames for each clip. A3D \cite{DBLP:conf/iros/YaoXWCA19} and DoTA \cite{yao2022dota} are used for unsupervised ego-view accident detection \cite{DBLP:conf/iros/YaoXWCA19,yao2022dota,fang2022traffic}. Specially, the DADA-2000 dataset \cite{DBLP:journals/tits/FangYQXY22} annotates the extra driver attention data. Because of the difficulty to collect enough accident videos in real world, some work leverage the simulation tool to synthesize the virtual accident videos or object tracklets, such as GTACrash \cite{DBLP:conf/aaai/KimLHS19}, VIENA$^2$\cite{aliakbarian2019viena}, DeepAccident \cite{DBLP:journals/corr/abs-2304-01168}, and CTAD \cite{luoICASSP2023}. However, the real-synthetic data domain gap is a tough nut to crack because it is rather hard to project the natural evolution process of accidents in the simulation tools. Besides CTA \cite{DBLP:conf/eccv/YouH20}, the vision modal is concentrated, and the meaningful text descriptions are not explored.
  
      \begin{table}[!t]\scriptsize
  \centering
  \caption{Static attributes of ego-view accident video datasets.}
     \setlength{\tabcolsep}{0.3mm}{
\begin{tabular}{c|cccc|ccccc}
    \toprule
 \multirow{2}[2]{*}{Datasets} & \multicolumn{4}{c|}{weather condition} & \multicolumn{5}{c}{occasion situations}\\
\cmidrule{2-10} & sunny &rainy&snowy& foggy &highway&urban &rural&mountain& tunnel \\
 \hline
CCD \cite{bao2020uncertainty} &1,306& 61 & 14&0&148& 725 & 502&5&1\\
A3D \cite{DBLP:conf/iros/YaoXWCA19} & 2,990& 251 &474&42& 225& 2,458 &720&328&26\\
DADA-2000 \cite{DBLP:journals/tits/FangYQXY22} & 1,860& 130 &10&-& \textbf{1,420}& 380 &180&-&20\\
DoTA \cite{yao2022dota} & 4,920& 341 &313&12& 617& 3,656 &1,148&145&20\\
\datasetname & \textbf{10,116}& \textbf{761}&\textbf{793}&\textbf{57}& 1082& \textbf{7,563}&\textbf{2,548}&\textbf{484}&\textbf{50}\\
    \toprule
  \end{tabular}}
  \label{tab2}
  \end{table}
  
  \begin{figure}[!t]
  \centering
\includegraphics[width=\linewidth]{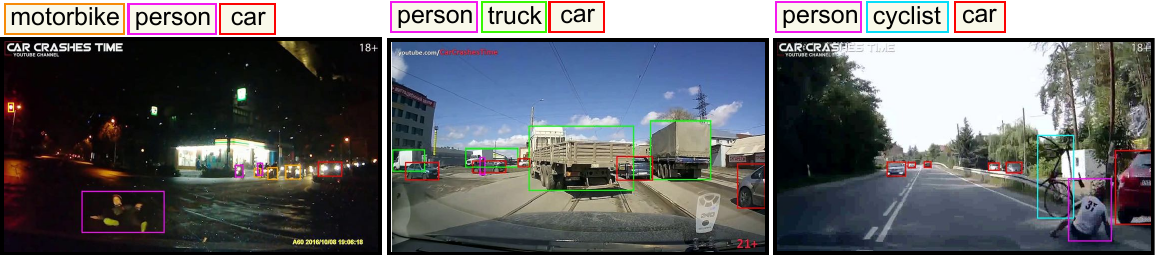}
   \caption{Some samples of object annotation in \datasetname.}
   \label{fig2}
   \vspace{-1.5em}
\end{figure}

  \begin{figure*}[!t]
  \centering
\includegraphics[width=\linewidth]{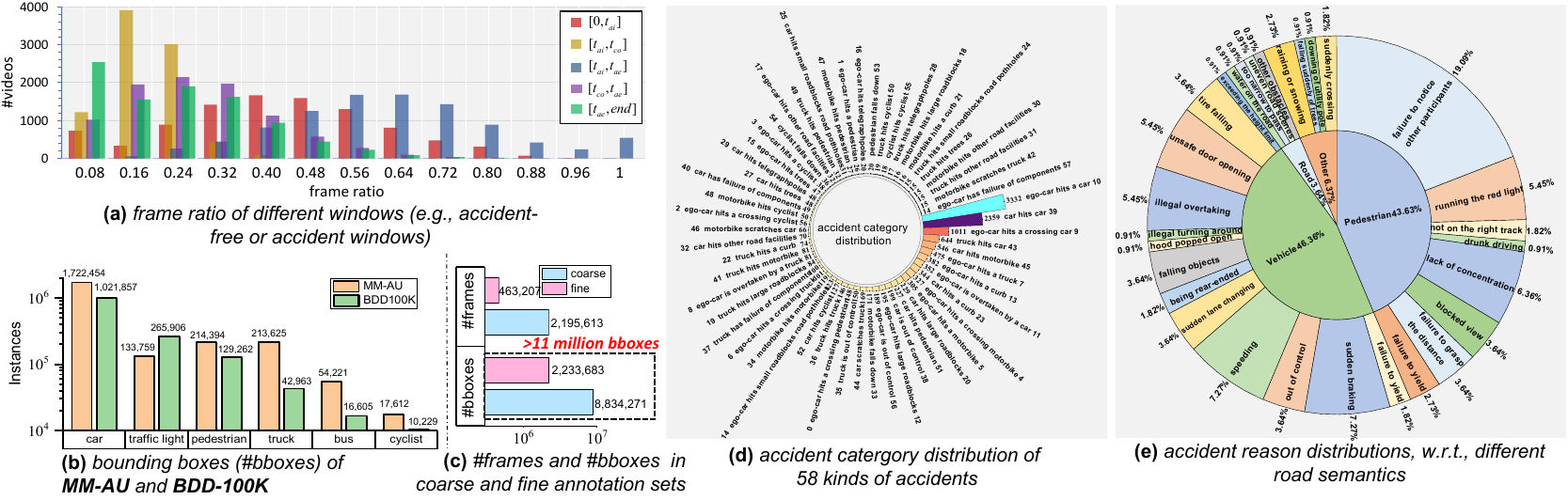}
   \caption{The annotation attribute statistics in \datasetname~ for the temporal, object, and text annotations. Better viewed in zoomed-in mode.}
   \label{fig3}
   \vspace{-0.5em}
\end{figure*}

\section{\datasetname~Dataset}
The videos in \datasetname~are collected from the publicly available ego-view accident datasets, such as CCD \cite{bao2020uncertainty}, A3D \cite{DBLP:conf/iros/YaoXWCA19}, DoTA \cite{yao2022dota}, and DADA-2000 \cite{DBLP:journals/tits/FangYQXY22}, and various video stream sites, such as Youtube \footnote{https://www.youtube.com}, Bilibili \footnote{https://www.bilibili.com}, and Tencent \footnote{https://v.qq.com}, \emph{etc}. As presented in Tab. \ref{tab2}, the weather conditions and occasion situations are various and our \datasetname~ owns the largest sample scale. In total, \textbf{11,727} videos with \textbf{2,195,613} frames are collected and annotated. All videos are annotated the text descriptions, accident windows, and accident time stamps. To the best of our knowledge, \datasetname~is the largest and most fine-grained ego-view multi-modal accident dataset. The annotation process of \datasetname~ is depicted as follows.

\textbf{Accident Window Annotation}: Leveraging the annotation criteria of DoTA \cite{yao2022dota}, the accident window is labeled by 5 volunteers, and the final frame indexes are determined by average operation. The temporal annotation contains the beginning time of the accident $t_{ai}$, the end time of the accident $t_{ae}$, and the beginning time of the collision $t_{co}$. The frame ratio distributions within different windows of [0,$t_{ai}$], [$t_{ai}$,$t_{co}$], [$t_{ai}$,$t_{ae}$], [$t_{co}$,$t_{ae}$], and [$t_{ae}$, end] are shown in Fig. \ref{fig3}(a). It can be seen that, in many videos, many accidents end in the last frame. The accident window of [$t_{ai}$,$t_{ae}$] of most videos occupies half of the video length, which is useful for model training in accident video understanding.

\textbf{Object Detection Annotation:} To facilitate the object-centric accident video understanding, we annotate 7 classes of road participants (\emph{i.e., cars, traffic lights, pedestrians, trucks, buses, and cyclists}) in \datasetname. To fulfill an efficient annotation, we firstly employ the YOLOX \cite{DBLP:journals/corr/abs-2107-08430} detector (pre-trained on the COCO dataset \cite{lin2014microsoft}), to initially detect the objects in the raw \datasetname~ videos. Secondly, \datasetname~ has two sets of bounding-box (\#bboxes) annotations. We name them as fine annotation set and coarse annotation set. For the fine annotation set, we took three months to manually correct the wrong detections using LabelImg every five frames by ten volunteers, and \textbf{2,233,683} bounding boxes within 463,207 frames are obtained. Each bounding box is double-checked for the final confirmation. Fig. \ref{fig2} presents some samples of object annotations in \datasetname. As for the coarse annotation set, we utilize the state-of-the-art (SOTA) DiffusionDet \cite{chen2023diffusiondet}\footnote{DiffusionDet is experimentally compared with 11 state-of-the-art detectors in Tab. \ref{tab3} after fine-tuning it on the fine annotation set.} to obtain the object-bounding boxes for the remainder of frames in \datasetname. Fig. \ref{fig2}(b) presents the \#bboxes on different road participants with the comparison to BDD-100K \cite{DBLP:conf/cvpr/YuCWXCLMD20}, and Fig. \ref{fig3}(c) shows the \#frames and \#bboxes on the fine and coarse annotation sets.

\textbf{Text Description Annotation:} Different from previous ego-view accident video datasets, \datasetname~ annotates three kinds of text descriptions: accident reason, prevention advice, and accident category descriptions. Accident category description is certainly aligned to the accident window $[t_{ai},t_{ae}]$, while the reason and prevention advice description are aligned to the near-accident window $[t_{ai}-40,t_{ai}]$. The descriptions and the video sequences do not show a unique correlation, and each description sentence usually correlates with many videos because of the co-occurrence. Similar to the work \cite{DBLP:conf/itsc/FangYQXWL19}, based on the road layout, road user categories, and their movement actions, we annotate $58$ description sentences for accident categories, and their sample distribution is shown in Fig. \ref{fig3}(d). 

We annotate $110$ pairs of sentences for accident reason and prevention advice descriptions correlating to four kinds of road semantics, \emph{i.e.}, pedestrian-centric, vehicle-centric, road-centric, and others (environmental issue). Fig. \ref{fig3}(e) shows the accident reason distribution concerning different road semantics. It is clear that the ``\emph{failure to notice other participants}" is dominant for pedestrian-centric accident reasons, and ``\emph{speeding}", ``\emph{sudden braking}", and ``\emph{illegal overtaking}" are the main kinds of vehicle-centric accident reasons. Following the form of Video Question Answering (VideoQA) task \cite{DBLP:journals/pami/XiaoZYLHYC23}, we provide an Accident reason Answering (ArA) task while there is only one question ``\texttt{What is the reason for the accident in this video?}". For each accident reason of a video, we further provide four reasonable distractors\footnote{(1) general distractor: distracted driving, speeding, extreme weather, etc. (2) location distractor: sudden overtaking or lane-changing in the main road, too fast in turning, running the red light in intersection, etc. (3) keyword distractor: motorcycle$\rightarrow$ cyclist, car$\rightarrow$ ego-car, decelerate$\rightarrow$ accelerate, stop$\rightarrow$ start, etc. (4) random distractor: random select one reason description in the reason set.} to form a multi-choice ArA task, and the distractor reasons are all unrelated to the target accident video. We obtain \textbf{58,650} ArA pairs in \datasetname.

\section{AdVersa-SD}
This section presents our AdVersa-SD with the abductive text-video coherent learning for ego-view accident video understanding. As aforementioned, we partition each accident video into three video segments, \emph{i.e.}, the normal video segment \textcolor{earthyellow}{$V_o$} \YellowRect, the near-accident segment \textcolor{highblue}{$V_{r}$} \BlueRect, and the accident segment \textcolor{magenta}{$V_a$} \PurpleRect. Correspondingly, we annotate the text descriptions of the accident reason \textcolor{highblue}{$t_r$}, the prevention advice \textcolor{highblue}{$t_p$}~, and the accident category \textcolor{magenta}{$t_a$}. To be clear, we define a denotation of text-video Co-oCcurrence Pair (\textbf{Co-CP}) to represent the natural co-occurrence of video clip and text description, \eg, (\textcolor{highblue}{$V_r$}, \textcolor{highblue}{$t_r$}), and (\textcolor{magenta}{$V_a$}, \textcolor{magenta}{$t_a$}). 
  \begin{figure}[!t]
  \centering
\includegraphics[width=0.95\linewidth]{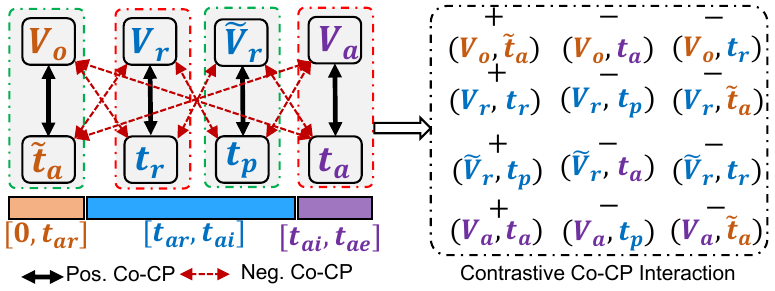}
   \caption{The structure of \textbf{Abductive CLIP} contains four interaction groups with one positive Co-CP and two negative Co-CPs for each interaction group, where $t_{ar}$=$t_{ai}$-40.}
   \label{fig4}\vspace{-0.8em}
\end{figure}

\subsection{Abductive CLIP} For abductive video understanding, we propose an Abductive CLIP for AdVersa-SD to fulfill the coherent semantic learning within different Co-CPs. The structure of Abductive CLIP is illustrated in Fig. \ref{fig4}. We create two virtual Co-CPs for training Abductive CLIP, \emph{i.e.}, the (\textcolor{brown}{$V_{o}$}, \textcolor{brown}{$\tilde{t}_{a}$}) and (\textcolor{highblue}{$\tilde{V}_{r}$}, \textcolor{highblue}{$t_p$}). (\textcolor{brown}{$V_{o}$}, \textcolor{brown}{$\tilde{t}_{a}$}) represents the Co-CP of antonymous accident category description \textcolor{brown}{$\tilde{t}_{a}$} and normal video clip \textcolor{brown}{$V_o$}, while (\textcolor{highblue}{$\tilde{V}_{r}$}, \textcolor{highblue}{$t_p$}) denotes the Co-CP of \textcolor{highblue}{$\tilde{V}_{r}$} and the prevention advice description \textcolor{highblue}{$t_p$}. Notably, \textcolor{brown}{$\tilde{t}_{a}$} is obtained by adding the antonym of verbs to the accident category description, such as ``does/do not", ``are not", etc. In addition, to make a dissipation process from the near-accident state to the normal state, \textcolor{highblue}{$\tilde{V}_{r}$} is obtained by reverse frame rearrangement of \textcolor{highblue}{$V_{r}$}.

Certainly, because each video segment may have different numbers of video frames, we create Co-CPs by coupling the text description with the randomly selected 16 successive frames within the certain video segment. Consequently, the training for Abductive CLIP is straightforward by enhanced difference learning of the embeddings of Co-CPs. 

\textbf{Contrastive Interaction Loss:} Abductive CLIP takes the XCLIP model \cite{ni2022expanding} as the backbone. Subsequently, we input each Co-CP into XCLIP to obtain the feature embedding of the video clip feature $\textbf{z}_v$ and the text feature $\textbf{z}_t$. To achieve the purpose stated in Fig. \ref{fig4}, we provide a Contrastive Interaction Loss (CILoss) to make the interactive Co-CP learning. The CILoss for different interaction groups of Co-CPs is the same, and consistently defined as:
\begin{equation}\small
\begin{aligned}
\mathcal{L}_{\text{CILoss}}=-\sum_{i=1}^{B}\log \frac{E({\textbf{z}_{v_i}^p,\textbf{z}_{t_i}^p)}}{\mathcal{K}} \verb'       '\\\vspace{-1.5em}
\mathcal{K}=\sum_{j=1}^{B} [E({\textbf{z}_{v_i}^p},\textbf{z}_{t_j}^p)+E({\textbf{z}_{v_i}^p},\textbf{z}^{n_1}_{t_{j\neq i}})+E({\textbf{z}_{v_i}^p,\textbf{z}^{n_2}_{t_{j\neq i}}})],
\end{aligned}
\vspace{-1em}
\end{equation}
where $E(\textbf{z}_v,\textbf{z}_t)=e^{{\textbf{z}_{v}}^{T}\textbf{z}_{t}/\tau}$ computes the coherence degree of video clip feature $\textbf{z}_v$ and text feature $\textbf{z}_t$. $B$ denotes the batchsize scale, the upscripts $p$, and $n_1/n_2$
refer to the \emph{Pos. CoCP} and the \emph{Neg. CoCPs}. $\tau$ is a learnable hyper-parameter, $i$ and $j$ are the sample indexes in each batchsize.

CILoss aims to enhance the coherence between the text description and video frames in Co-CP by enlarging the distance of text descriptions or video frames with the ones in negative Co-CPs. Abductive CLIP is optimized by minimizing the summation of four kinds of $\mathcal{L}^{o, r, p, a}_{\text{CILoss}}$.

\subsection{Extension to Accident Video Diffusion} 

\begin{figure}[!t]
  \centering
\includegraphics[width=\linewidth]{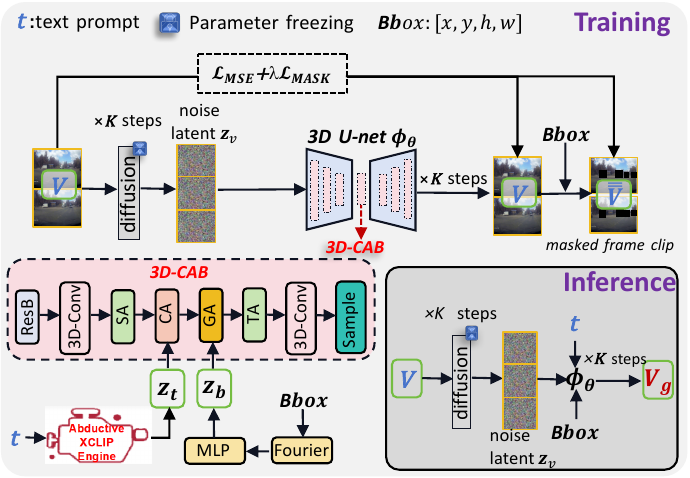}
   \caption{The structure of the object-centric Accident Video Diffusion model (OAVD), where the $V$ and $t$ are the video frame clip and the corresponding text prompt. Object bounding boxes $Bbox$ in each video frame that can be obtained by object detectors. $\overline{\overline{V}}$ is the masked frame clip where the pixels in object regions are set to 0. Different attention modules, \emph{i.e.}, SA, CA, and TA, follow the ones of \cite{wu2023tune} while with a dense multi-head attention.}
   \label{fig5}\vspace{-1.5em}
\end{figure}

To verify Abductive CLIP, this work treats it as an engine to drive the accident video diffusion task for explicitly exploring the causal-effect relation of accident occurrence. Because traffic accidents are commonly caused by the irregular or sudden movement of road participants, the video diffusion model should have the ability for object-level representation. As shown in Fig. \ref{fig5}, we propose an Object-centric Accident Video Diffusion model (OAVD) which takes the Latent Diffusion Model (LDM) \cite{rombach2022high} as the baseline and extends it to the video diffusion with the input of Co-CPs and $K$ steps of forward and reverse diffusion process. 

The structure of OAVD becomes similar to Tune-A-Video work \cite{wu2023tune}, while differently, the 3D-CAB block of the 3D U-net module in Fig. \ref{fig5} is redesigned and contains the 3D-Conv layers (with four layers with the kernel size of (3,1,1)), a text-video Cross-Attention (\textbf{CA}) layer, a Spatial Attention (\textbf{SA}) layer, a Temporal Attention (\textbf{TA}) layer and a Gated self-attention (\textbf{GA}) \cite{li2023gligen} for the frame correlation and object location consideration. To be capable of object-level video diffusion, We further add a masked representation reconstruction path on frame-level reconstruction. The \textbf{inference} phase has the same input form as the OAVD training and generates the new frame clip $V_g$.

\textbf{Masked Video Frame Diffusion:} Our aim is to learn the 3D U-net by forward adding noise on the clean latent representation of raw video frames and inverse denoising the noise $\textbf{z}_v$ with $K$ time steps, conditioned by the text prompt $t$ and the bounding boxes. This formulation enables the derivation of the mask video representation $\textbf{z}_{mask}$ of $\overline{\overline{V}}$ to fix the frame background details in diffusion process and fulfill the object-centric video generation. The optimization of 3D U-net is achieved by minimizing: 
\begin{equation}
\begin{aligned}
\mathbb{E}\substack{V,\textbf{e} \sim \substack{\mathcal{N}(0,I)},k,\textbf{z}_t,\textbf{z}_b,\overline{\overline{V}}}\Bigl\{ \lVert \textbf{e} - \phi_{\theta}(\textbf{z}_v, k, \textbf{z}_t,\textbf{z}_b) \rVert_2^2 + \\ 
\lambda \lVert \textbf{e}(\textbf{1}-\textbf{z}_{mask}) - \phi_{\theta}(\textbf{z}_v, k, \textbf{z}_t,\textbf{z}_b)(\textbf{1}-\textbf{z}_{mask}) \rVert_1^1 \Bigr\},
\end{aligned}
\end{equation}
where the first term is Mean Square Error ($\mathcal{L}_{MSE}$) and the second term denotes the reconstruction loss of the Masked Latent Representation ($\mathcal{L}_{MASK}$). $k\in[1,\dots, K]$ denotes the diffusion step ($K$=1000), $\textbf{e}$ is the ground-truth noise representation in each diffusion step. $\phi_{\theta}$ is the 3D U-net to be optimized which contains the parameters of 3D Cross-Attention Blocks (\textbf{3D-CAB}) with down-sample and up-sample layers (\textbf{Sample}). $\lambda$=$0.5$ is a parameter for balancing the weights of $\mathcal{L}_{MSE}$ and $\mathcal{L}_{MASK}$. $\textbf{1}$ is an identity tensor with the same size of $\textbf{e}$, and the masked noise representation $\textbf{z}_{mask}$ is obtained by Denoising Diffusion Probabilistic Model (DDPM) Scheduler \cite{ho2020denoising} on the binarization of a latent representation $\textbf{z}_{l}$ ($\textbf{z}_{l}=\text{VAE}(\textbf{z}_{v})$) through the Variational Autoencoder (VAE)in LDM \cite{rombach2022high} as:
\begin{equation}
\begin{aligned}
\textbf{z}_{mask} = \text{DDPM Scheduler}(\textbf{m}_{(z)},k,\textbf{e})),\\
\textbf{m}_{(z)} =
  \begin{cases} 
      0 & \text{if } \textbf{z}_{l} < 0.5, \\
      1 & \text{if } \textbf{z}_{l}  \geq 0.5. \verb'          '
  \end{cases}
\end{aligned}
\end{equation}

\textbf{Gated Bbox Representation:} The key insight is to involve object bounding boxes to enhance the causal object region learning concerning the related text words, which is useful for eliminating the influence of the frame background and explicitly checking the role of certain text words for subsequent accident situations. Inspired by the Gated self-Attention (GA) \cite{li2023gligen}, the location embedding $\textbf{z}_b$,  collaborating with the output of CA in 3D-CAB, is obtained by:
\begin{equation}
\textbf{z}_b=\text{MLP}(\text{Fourier}(Bbox)),
\label{eq:Fourier}
\end{equation}
where $\textbf{z}_b$ is obtained from Bbox by MLP layers with the Fourier embedding \cite{mildenhall2021nerf}.

\textbf{OAVD Inference:} The OAVD inference stage inputs the Co-CPs while Denosing Diffusion Implict Model (DDIM) scheduler \cite{DBLP:conf/iclr/SongME21} is taken on the trained 3D U-net $\phi_\theta$ conditioned by the text prompt $t$ and $Bbox$. The frame clip $V$ within the Co-CPs are fed into the inference stage with the same dimension to the generated $V_g$. 

\begin{table*}[!t]
\scriptsize
\centering
\caption{\small{The results of V1-Train [\YellowRect, \BlueRect, \PurpleRect]) and V2-Train [\YellowRect, \BlueRect]) for 11 state-of-the-art detectors on the \datasetname.}}
\setlength{\tabcolsep}{1.6mm}{
\begin{tabular}{c|c|cc|cc|cc|cc|cc|c|c|c}
    \toprule
\multirow{4}{*}{Detectors} & \multirow{4}{*}{Years} & \multicolumn{6}{c|}{V1-Train [\YellowRect, \BlueRect, \PurpleRect]} & \multicolumn{4}{c|}{V2-Train [\YellowRect, \BlueRect]} & \multirow{4}{*}{\makebox[0.35cm][c]{Anchor}} & \multirow{4}{*}{GFlops} & \multirow{4}{*}{\#Params.} \\
\cline{3-12}
                           &                        & \multicolumn{2}{c|}{val. \YellowRect, \BlueRect, \PurpleRect} & \multicolumn{2}{c|}{test. \YellowRect, \BlueRect, \PurpleRect} & \multicolumn{2}{c|}{test. \PurpleRect} & \multicolumn{2}{c|}{test. \YellowRect, \BlueRect} & \multicolumn{2}{c|}{test. \PurpleRect} & & & \\
\cline{3-12}
                           &                        & mAP50& AR & mAP50& AR & mAP50 & AR & mAP50 &AR &mAP50 & AR & & & \\ \hline
FasterRCNN \cite{ren2015faster}               & 2015                   & 0.674&0.634&0.666&0.623 &0.664  & 0.620   &0.544 & 0.524 & 0.497  & 0.509  & \checkmark & 0.19T                  & 41.38M                    \\ 
CornerNet \cite{law2018cornernet}                  & 2018                   &0.495&0.625 & 0.485 & 0.619 & 0.483 & 0.624  &0.436&0.563 & 0.456 & 0.598  &  & 0.71T                  & 201M                       \\ 
CascadeRPN \cite{vu2019cascade}                & 2019                   &0.662&0.699& 0.664 & 0.696 &0.649  & 0.689  & 0.579& 0.663 &0.532 & 0.624  & \checkmark & 0.18T                  & 41.97M                    \\ 
CenterNet \cite{duan2019centernet}                  & 2019                   &0.054 & 0.238& 0.051 & 0.233 & 0.047 & 0.224  & 0.161 &0.260  &0.155 & 0.257  &  & 20.38G                 & 14.21M                    \\ 
DeTR \cite{detr2020}                      & 2020                   &0.367 &0.407 &0.377&0.403 &0.363 & 0.403  &0.275 & 0.329 & 0.254  & 0.318  &  & 44.55G                 & 28.83M                    \\
EfficientNet \cite{DBLP:conf/icml/TanL19}                & 2020                   & 0.310  & 0.417 & 0.310  & 0.412& 0.293  & 0.404  & 0.075& 0.128 & 0.073  & 0.133  &  & 57.28G                 & 18.46M                    \\ 
Deformable-DeTR \cite{zhu2021deformable}             & 2021                   & 0.660  &0.671 & 0.661 &0.668& 0.652  & 0.663  & 0.626 & 0.631 & 0.587  & 0.626  &  & 0.18T                  & 40.1M                      \\
YOLOx \cite{DBLP:journals/corr/abs-2107-08430}                      & 2021                   & 0.673& 0.709 & 0.672 & 0.698& 0.670   & 0.698  & 0.563 & 0.627 &0.540  & 0.626  &  & 13.33G                 & 8.94M                      \\ 
YOLOv5s \cite{glenn_jocher_2022_7002879}                     & 2021                   & \cellcolor{LightCyan}\textbf{0.757} & \cellcolor{LightCyan}\textbf{0.766} & \cellcolor{LightCyan}\textbf{0.748} & \cellcolor{LightCyan}\textbf{0.764} & \cellcolor{LightCyan}\textbf{0.743}  &\cellcolor{LightCyan}\textbf{0.761}  & 0.660  & 0.716 & 0.636 & 0.712  & \checkmark & 8.13G                  & 12.35M                    \\ 
DiffusionDet  \cite{chen2023diffusiondet}             & 2023                   & 0.731 & 0.749 &0.733 & 0.745 &0.718  & 0.738  & \cellcolor{LightCyan}\textbf{0.701}& \cellcolor{LightCyan}\textbf{0.729} &\cellcolor{LightCyan}\textbf{0.660}   & \cellcolor{LightCyan}\textbf{0.716}  &  & -                       & 26.82M                    \\ 
YOLOv8 \cite{ultralyticsgithub}                      & 2023                   & 0.716& 0.754 & 0.715& 0.753&0.717  & 0.755  & 0.606 & 0.702 & 0.597  & 0.703  &  & 14.28G                 & 11.14M                    \\     \toprule
\end{tabular}}
\begin{tablenotes}
\item \scriptsize{\textbf{Notes}: To ensure that the distribution of accident windows is essentially the same on the training, validation, and testing sets, we divide the bounding box annotations in the ratio of 7:1.5:1.5. The results of GFlops and \#Params. are reported by the tools in MMDetection and MMYOLO of the testing phase.}
\end{tablenotes}
\label{tab3}\vspace{-0.5em}
\end{table*}

\section{Experiments}
\subsection{Experimental Details}
AdVersa-SD takes as input the object bounding boxes and accident reasons. As a prerequisite, we first carry out a benchmark evaluation for the Object Detection (\textbf{OD}) and Accident reason Answering (\textbf{ArA}) tasks, which is of crucial importance to video diffusion. Second, we evaluate our \textbf{AdVersa-SD} with extensive video diffusion experiments.

\textbf{(1) OD Task}: We select 11 state-of-the-art detectors to be presented in Tab. \ref{tab3} in the OD benchmark evaluation.
All detectors used the corresponding architectures provided in MMDetection \cite{mmdetection} and MMYOLO \footnote{https://github.com/open-mmlab/mmyolo} while keeping important hyperparameters equal, such as batch size, initial learning rate, and epochs. All training and inference are implemented on three GeForce RTX 3090s. All the detectors are pre-trained on the BDD-100K dataset \cite{DBLP:conf/cvpr/YuCWXCLMD20}, and fine-tuned with the training set of fine object annotations. Notably, to check the OD performance in accident window \PurpleRect, we provide two versions of detectors fine-tuned on the training set coming from whole frame windows (abbrev., V1-Train [\YellowRect, \BlueRect, \PurpleRect]) and the ones fine-tuned on the training set of accident-free windows (abbrev., V2-Train [\YellowRect, \BlueRect]).

We use Average Precision (\textbf{AP50}) and Average Recall (\textbf{AR}) \cite{lin2014microsoft} to evaluate the detection results with the threshold of 50\% detection score.

\textbf{(2) ArA Task:} We follow the task of multi-choice Video Question Answering (VQA) to formulate the ArA task while the question is ``\texttt{What is the reason for the accident in this video?}". The performance is measured by the \textbf{accuracy}, \emph{i.e.}, the percentage of questions that are correctly answered.

\textbf{(3) Abductive Video Diffusion Task:} 
In AdVersa-SD, we aim to explore the cause-effect evolution of accident videos conditioned by the descriptions of accident reasons or prevention advice. Hence, based on the input form of OAVD in AdVersa-SD, we input the object bboxes and Co-CPs of (\textcolor{highblue}{$V_{r}$}, \textcolor{highblue}{$t_r$}) and (\textcolor{highblue}{$V_{r}$}, \textcolor{highblue}{$t_p$}) in the evaluations. Two state-of-the-art video diffusion models including the DDIM inversion-based Tune-A-Video (TAV) \cite{wu2023tune} and the training-free ControlVideo (CVideo) \cite{zhang2023controlvideo} are selected. We generate 1500 clips (with 16 frames/clip) for all diffusion experiments, in which the object boxes are pre-detected by DiffusionDet~\cite{chen2023diffusiondet}. 

Similar to previous video diffusion models, Fréchet Video Distance (\textbf{FVD}) \cite{unterthiner2018towards} is taken for quality evaluation of the synthetic video clips. We also use the CLIP score (\textbf{CLIP}$_S$) \cite{wu2023tune} to measure the alignment degree between text prompts and video frames.

In the evaluation, 6000 pairs of Co-CPs in \datasetname~are adopted to train the AdVersa-SD and the Tune-A-Video model. The learning rate of Abductive CLIP in AdVersa-SD is $1e-6$ with the batchsize of 2 and trained with 30000 iteration steps. The learning rate of OAVD is $5e-6$ with the same batchsize and trained with 8000 iteration steps. Adam optimizer is adopted with the default $\beta_1=0.9$ and $\beta_2=0.999$ on a platform with 2 GeForce RTX 3090s.

\subsection{Result Analysis}

\textbf{(1) OD Evaluations:} Tab. \ref{tab3} presents the detection results of 11 state-of-the-art detectors. From the results, we can see all the detectors generate a degradation for the accident window test. CenterNet \cite{duan2019centernet} and EfficientNet \cite{DBLP:conf/icml/TanL19} show limited ability for the OD task in the accident scenarios and the metric values decrease significantly for V2-Train mode. As claimed by previous research, pure Transformer-based detectors, such as DeTR \cite{detr2020}, demonstrate limited performance. Deformable-DeTR has improved performance but is still not better than the CNN-based ones for traffic accident cases. YOLOv5s \cite{glenn_jocher_2022_7002879} and DiffusionDet~\cite{chen2023diffusiondet} are the two leading approaches for the OD task. However, from the results difference obtained by V1-Train [\YellowRect, \BlueRect, \PurpleRect] and V2-Train [\YellowRect, \BlueRect], DiffusionDet \cite{chen2023diffusiondet} shows superior performance to the testing set of accident window \PurpleRect~in V2-Train. It indicates that diffusion-based object detection may be more robust with better generalization ability. More qualitative results can be viewed in the supplemental file.

\begin{figure}[!t]
  \centering
\includegraphics[width=0.98\linewidth]{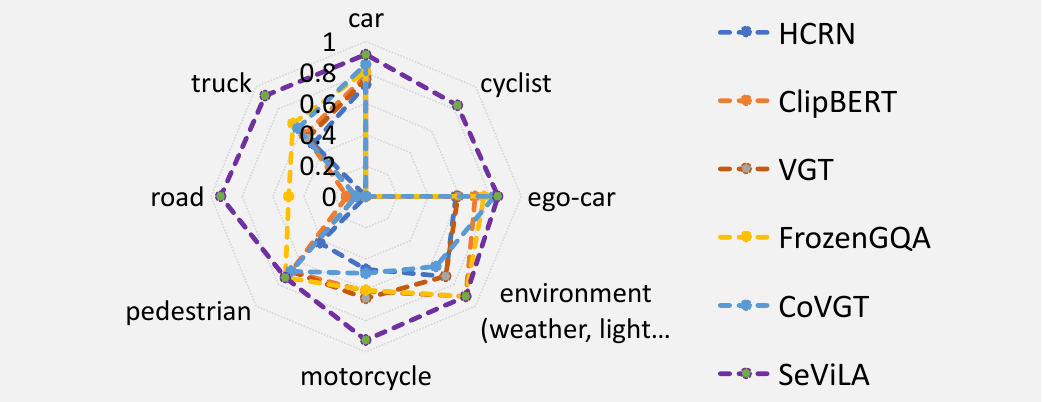}
   \caption{\small{Accident Reason Answering (ArA) Accuracy (Acc. \%), w.r.t., different accident participants on the testing set of \datasetname.}}
   \label{fig6}
   \vspace{-1em}
\end{figure}

\textbf{(2) ArA Evaluations:}
We present in Tab.~\ref{tab:ara_res} and Fig.~\ref{fig6} the performances of the state-of-the-art on the ArA task. We carefully select the baseline methods to include temporal relation network (HCRN~\cite{le2020hierarchical}), graph transformer network (VGT~\cite{xiao2022video}, CoVGT~\cite{DBLP:journals/pami/XiaoZYLHYC23}), cross-modal pre-trained transformers (ClipBERT~\cite{lei2021less}) and those using large language models (LMMs) (FrozenGQA~\cite{xiao2023can} and SeViLA~\cite{yu2023self}). The methods also include frame-centric and more fine-grained object-centric video representations. Our key observations are: LLM-based methods, such as SeViLA which uses Flan T5-XL (3B)~\cite{chung2022scaling}, show absolute advantage in this task, surpassing the second-ranked method CoVGT by 7\% to 9\%. Furthermore, fine-grained visual representations, \eg, region or object level, are key for higher performances. We speculate that the videos are all taken on the road about traffic accidents and thus a coarse frame-level representation is insufficient to discern various accident reasons. 
\begin{table}[!t]\small
\centering
\caption{\small{The Accident Reason Answering (ArA) Accuracy (Acc. \%) on the validation (val.) and testing (test.) set of \datasetname~ by 6 SOTA methods whose size of learnable parameters is provided.}}
     \setlength{\tabcolsep}{1mm}{
     \scalebox{0.9}{
\begin{tabular}{l|c|c|c|c|c|c}
    \toprule
Methods & Years &Acc (val.)      & Acc (test.)     & V. & T. &Params.(M) \\ \hline
HCRN \cite{le2020hierarchical}                     & 2020                   & 65.81   & 64.65        &    F   & G              &     42                    \\ 
ClipBERT \cite{lei2021less}                & 2021                   & 72.09    & 72.71      &    F      & B             &   137                    \\ 
VGT \cite{xiao2022video}                   & 2022                   & 68.40    & 68.66       &   O       & B             &         143                \\ 
FrozenGQA \cite{xiao2023can}                & 2023                   & 77.10    & 77.01       &     F       & D           &       30                 \\ 
CoVGT \cite{DBLP:journals/pami/XiaoZYLHYC23}  & 2023    &81.70   & 79.97      &     O       & R           &                         159\\ 
SeViLA \cite{yu2023self}                  & 2023                   & \textbf{89.26}    & \textbf{89.02}    &  O & F   &       108                 \\     \toprule
\end{tabular}
}}
    \begin{tablenotes} 
\item \scriptsize{F: frame-centric representations; O: object-centric representations; V.:Vision; T.:Text; G: GloVe; B: BERT \cite{devlin2018bert}; D: DeBERTa \cite{he2020deberta}; R: RoBERTa \cite{liu2019roberta}; F: Flan T5 \cite{chung2022scaling}. The ratio of training, validation, and testing set is 7:1:2.}
\end{tablenotes}
\vspace{-0.5em}
\label{tab:ara_res}
\end{table}

  \begin{figure*}[!t]
  \centering
\includegraphics[width=0.95\linewidth]{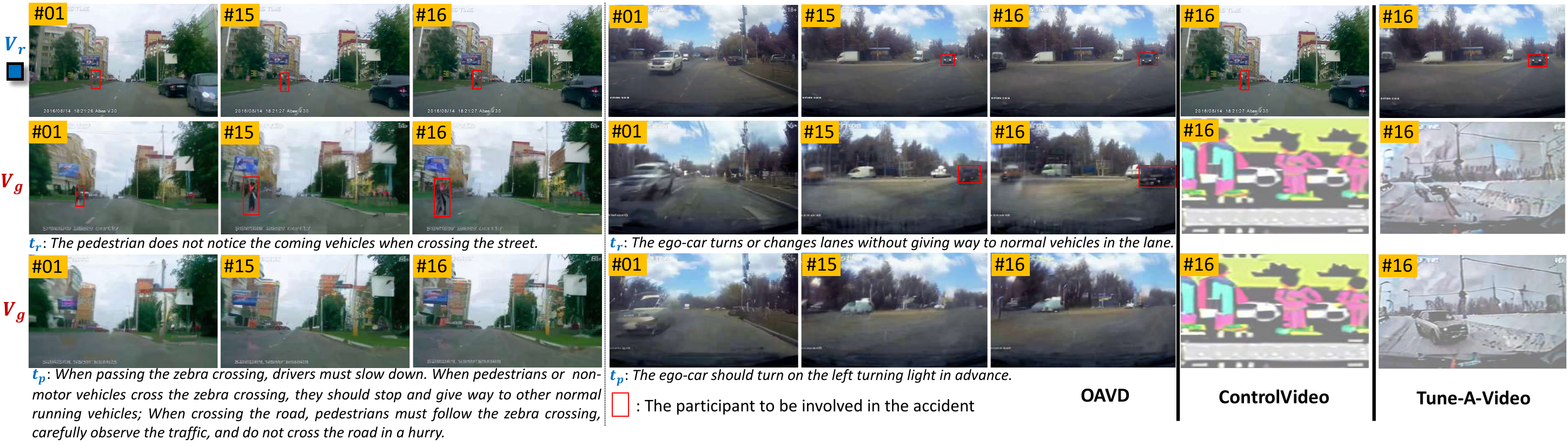}
   \caption{\small{Some results (\textcolor{red}{$V_g$}) inputting the Co-CPs of (\textcolor{highblue}{$V_{r}$}, \textcolor{highblue}{$t_r$}) or (\textcolor{highblue}{$V_{r}$}, \textcolor{highblue}{$t_p$}). From the generation of OAVD, the participants to be involved in accidents appear in advance when giving the accident reason prompt \textcolor{highblue}{$t_r$}, while the accident objects disappear when providing the prevention advice prompt \textcolor{highblue}{$t_p$}. ControlVideo and Tune-A-Video are given the same \textcolor{highblue}{$t_r$} and \textcolor{highblue}{$t_p$} prompt with OAVD, respectively. However, the artifacts and unrelated content are generated, and the phenomena of ``\emph{appear in advance} ” or ``\emph{disappear}” do not occur.}}
   \label{fig7}
   \vspace{-1em}
\end{figure*}

An in-depth analysis in Fig.~\ref{fig6} shows that the methods' performances vary a lot across different accident participant types. Generally, SeViLA outperforms other methods in all scenarios. Yet, all methods perform closely when the accident reason is car-related. Curiously, we find that most of the methods fail to identify the cyclists in the accidents except for SeViLA. The reason could be that cyclists are the least frequent participants (as shown in Fig. \ref{fig3} (b)) in accidents. Thus, it calls for commonsense knowledge (carried in LLMs) to find the related reasons.    

\textbf{(3) Diffusion Evaluations:} Here, we evaluate the AdVersa-SD, and the importance of Abductive-CLIP and Bboxes. To verify Abductive-CLIP in AdVersa-SD, we take two baselines: 1) the original CLIP model \cite{radford2021learning} and 2) a ``Sequential-CLIP (S-CLIP)" that only maintains the positive Co-CPs of the Abductive-CLIP (abbrev., A-CLIP) structure (see Fig. \ref{fig4}). 

 \noindent \textbf{Abductive Ability Check of AdVesrsa-SD}. Fig. \ref{fig7} visually presents video diffusion results of accident scenarios given the descriptions of accident reason \textcolor{highblue}{$t_{r}$} and prevention advice \textcolor{highblue}{$t_{p}$}, respectively. Curiously, OAVD can make the accident participant (\emph{i.e.}, the pedestrian or the black car) appear in advance given \textcolor{highblue}{$t_{r}$}, and eliminate the accident participants provided \textcolor{highblue}{$t_{p}$}. It indicates that our AdVesrsa-SD catches the dominant object representation for the cause-effect chain of the accident occurrence. Contrarily, ControlVideo and Tune-A-Video generate irrelevant styles with worse performance than OAVD, as listed in Tab. \ref{tab5}, which shows that accident knowledge is scarce in this field but rather crucial for the accident video diffusion models. Tab. \ref{tab5} shows the results of different diffusion models.

\noindent \textbf{Roles of different CLIP Models}. Tab. \ref{tab5} also presents the results of OAVD with varying CLIP models. The results show that our Abductive-CLIP can generate better text-video semantic alignment than the original CLIP model and the Sequential-CLIP trained on our MM-AU. It indicates that the contrastive interaction loss of the text-video pairs, \emph{i.e.}, Co-CPs, is important to discern the key semantic information within text and videos.
\begin{table}[!t]\scriptsize
  \centering
  \caption{\small{Results with SOTA diffusion models and our OAVD driven by varying CLIP models, where $\downarrow$ and $\uparrow$ prefer a lower and larger value, respectively. FPS: frames/second (tested on a single GeForce RTX 3090).}}
      \renewcommand{\arraystretch}{1.2}
 \setlength{\tabcolsep}{0.1mm}{
 \begin{tabular}{c|c|c|c|c|c}
     \toprule
Method & TAV \cite{wu2023tune}  & CVideo \cite{zhang2023controlvideo}  & OAVD (CLIP~\cite{radford2021learning}) & OAVD (S-CLIP)$^{*}$& OAVD (A-CLIP)$^{*}$\\
\hline
CLIP$_S$ $\uparrow$  & 21.77  & 22.51   & 21.9        & 27.14 &\textbf{27.24}\\
FVD $\downarrow$    & 9545.6 & 12275.2 & 10122.5     & 5372.3&\textbf{5238.1}\\
FPS $\uparrow$    & 1.7    & 0.5     & 1.7         & 1.2&1.2          \\     \toprule
\end{tabular}}
    \begin{tablenotes} 
\item \footnotesize{\textbf{*: with the input of bounding boxes.}}
\end{tablenotes}
\vspace{-0.5em}
\label{tab5}
\end{table}

\begin{figure}[!t]
  \centering
\includegraphics[width=\linewidth]{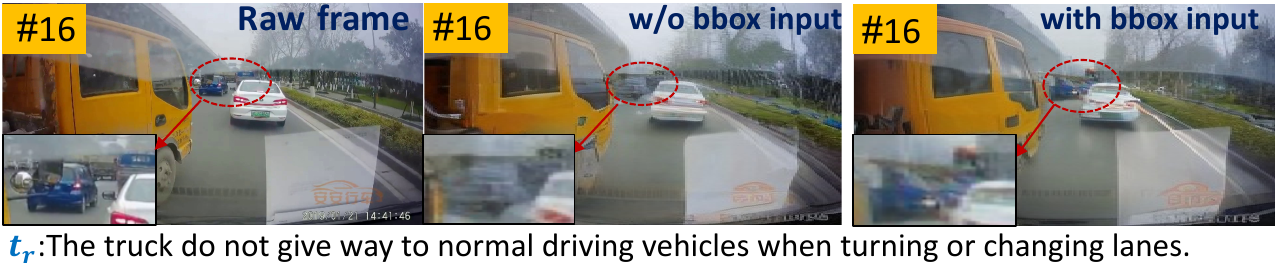}
   \caption{\small{A visualization for the importance of bounding box.}}
   \label{fig8}
   \vspace{-1em}
\end{figure}

 \noindent \textbf{Role of Bboxes} From Tab. \ref{tab5} and Fig. \ref{fig8}, the advantages of Bboxes are demonstrated with clearer and more detailed content in the generated frames. In addition, object-involved video diffusion can facilitate the key object region learning and maintain the details of the frames better than the version without Bbox input. More ablation studies on the role of bboxes can be viewed in the supplemental file.

Besides, our OAVD also can flexibly generate any accident videos with the input of only object boxes or the accident category descriptions (see the supplemental file).

\section{Conclusion}
This work presents a precious large-scale ego-view multi-modal accident dataset (\datasetname) for safe driving perception which provides the temporal, object, and text annotations for fine-grained accident video understanding. Within \datasetname, the evaluations of the state-of-the-art methods on object detection and accident reason answering tasks are carefully conducted. Based on \datasetname, we present AdVersa-SD to fulfill an abductive accident video understanding, where an Object-centric Accident Video Diffusion (OAVD) driven by an Abductive-CLIP model is proposed. Extensive experiments verify that AdVersa-SD shows promising ability for the cause-effect understanding of accident situations and generates superior video diffusion performance to two state-of-the-art diffusion models.

{\small
\bibliographystyle{ieee_fullname}
\bibliography{AAVD}

\begin{thebibliography}{10}\itemsep=-1pt

\bibitem{aliakbarian2019viena}
Mohammad~Sadegh Aliakbarian, Fatemeh~Sadat Saleh, Mathieu Salzmann, Basura Fernando, Lars Petersson, and Lars Andersson.
\newblock {VIENA}: A driving anticipation dataset.
\newblock In {\em ACCV}, pages 449--466, 2019.

\bibitem{bao2020uncertainty}
Wentao Bao, Qi Yu, and Yu Kong.
\newblock Uncertainty-based traffic accident anticipation with spatio-temporal relational learning.
\newblock In {\em ACM MM}, pages 2682--2690, 2020.

\bibitem{DBLP:conf/iccv/Bao0K21}
Wentao Bao, Qi Yu, and Yu Kong.
\newblock {DRIVE:} deep reinforced accident anticipation with visual explanation.
\newblock In {\em ICCV}, pages 7599--7608, 2021.

\bibitem{detr2020}
Nicolas Carion, Francisco Massa, Gabriel Synnaeve, Nicolas Usunier, Alexander Kirillov, and Sergey Zagoruyko.
\newblock End-to-end object detection with transformers.
\newblock In {\em ECCV}, pages 213--229, 2020.

\bibitem{DBLP:conf/accv/ChanCXS16}
Fu{-}Hsiang Chan, Yu{-}Ting Chen, Yu Xiang, and Min Sun.
\newblock Anticipating accidents in dashcam videos.
\newblock In {\em ACCV}, volume 10114, pages 136--153, 2016.

\bibitem{mmdetection}
Kai Chen et~al.
\newblock {MMDetection}: Open mmlab detection toolbox and benchmark.
\newblock {\em arXiv preprint arXiv:1906.07155}, 2019.

\bibitem{chen2023diffusiondet}
Shoufa Chen, Peize Sun, Yibing Song, and Ping Luo.
\newblock Diffusiondet: Diffusion model for object detection.
\newblock In {\em ICCV}, pages 19830--19843, 2023.

\bibitem{chung2022scaling}
Hyung~Won Chung, Le Hou, Shayne Longpre, Barret Zoph, Yi Tay, William Fedus, Yunxuan Li, Xuezhi Wang, Mostafa Dehghani, Siddhartha Brahma, et~al.
\newblock Scaling instruction-finetuned language models.
\newblock {\em arXiv preprint arXiv:2210.11416}, 2022.

\bibitem{devlin2018bert}
Jacob Devlin, Ming-Wei Chang, Kenton Lee, and Kristina Toutanova.
\newblock Bert: Pre-training of deep bidirectional transformers for language understanding.
\newblock {\em arXiv preprint arXiv:1810.04805}, 2018.

\bibitem{duan2019centernet}
Kaiwen Duan, Song Bai, Lingxi Xie, Honggang Qi, Qingming Huang, and Qi Tian.
\newblock Centernet: Keypoint triplets for object detection.
\newblock In {\em ICCV}, pages 6569--6578, 2019.

\bibitem{nature2022}
Editorial.
\newblock Safe driving cars.
\newblock {\em Nat. Mach. Intell.}, 4:95--96, 2022.

\bibitem{esser2023structure}
Patrick Esser, Johnathan Chiu, Parmida Atighehchian, Jonathan Granskog, and Anastasis Germanidis.
\newblock Structure and content-guided video synthesis with diffusion models.
\newblock In {\em ICCV}, pages 7346--7356, 2023.

\bibitem{fang2022traffic}
Jianwu Fang, Jiahuan Qiao, Jie Bai, Hongkai Yu, and Jianru Xue.
\newblock Traffic accident detection via self-supervised consistency learning in driving scenarios.
\newblock {\em {IEEE} Trans. Intell. Transp. Syst.}, 23(7):9601--9614, 2022.

\bibitem{DBLP:conf/itsc/FangYQXWL19}
Jianwu Fang, Dingxin Yan, Jiahuan Qiao, Jianru Xue, He Wang, and Sen Li.
\newblock {DADA-2000:} can driving accident be predicted by driver attention? {A}nalyzed by {A} benchmark.
\newblock In {\em ITSC}, pages 4303--4309, 2019.

\bibitem{DBLP:journals/tits/FangYQXY22}
Jianwu Fang, Dingxin Yan, Jiahuan Qiao, Jianru Xue, and Hongkai Yu.
\newblock {DADA:} driver attention prediction in driving accident scenarios.
\newblock {\em {IEEE} Trans. Intell. Transp. Syst.}, 23(6):4959--4971, 2022.

\bibitem{DBLP:journals/corr/abs-2107-08430}
Zheng Ge, Songtao Liu, Feng Wang, Zeming Li, and Jian Sun.
\newblock {YOLOX:} exceeding {YOLO} series in 2021.
\newblock {\em CoRR}, abs/2107.08430, 2021.

\bibitem{ghosh2019accident}
Sreyan Ghosh, Sherwin~Joseph Sunny, and Rohan Roney.
\newblock Accident detection using convolutional neural networks.
\newblock In {\em IconDSC}, pages 1--6, 2019.

\bibitem{DBLP:conf/avss/HajriF22}
Feten Hajri and Hajer Fradi.
\newblock Vision transformers for road accident detection from dashboard cameras.
\newblock In {\em AVSS}, pages 1--8, 2022.

\bibitem{he2020deberta}
Pengcheng He, Xiaodong Liu, Jianfeng Gao, and Weizhu Chen.
\newblock Deberta: Decoding-enhanced bert with disentangled attention.
\newblock {\em arXiv preprint arXiv:2006.03654}, 2020.

\bibitem{ho2020denoising}
Jonathan Ho, Ajay Jain, and Pieter Abbeel.
\newblock Denoising diffusion probabilistic models.
\newblock {\em NeurIPS}, 33:6840--6851, 2020.

\bibitem{DBLP:journals/eswa/HuWCG21}
Hongyu Hu, Qi Wang, Ming Cheng, and Zhenhai Gao.
\newblock Cost-sensitive semi-supervised deep learning to assess driving risk by application of naturalistic vehicle trajectories.
\newblock {\em Expert Syst. Appl.}, 178:115041, 2021.

\bibitem{DBLP:conf/icra/JainSKSS16}
Ashesh Jain, Avi Singh, Hema~Swetha Koppula, Shane Soh, and Ashutosh Saxena.
\newblock Recurrent neural networks for driver activity anticipation via sensory-fusion architecture.
\newblock In {\em ICRA}, pages 3118--3125, 2016.

\bibitem{glenn_jocher_2022_7002879}
Glenn Jocher et~al.
\newblock {ultralytics/yolov5: v6.2 - YOLOv5 Classification Models, Apple M1, Reproducibility, ClearML and Deci.ai integrations}, 2022.

\bibitem{kang2022vision}
Minhee Kang, Wooseop Lee, Keeyeon Hwang, and Young Yoon.
\newblock Vision transformer for detecting critical situations and extracting functional scenario for automated vehicle safety assessment.
\newblock {\em Sustainability}, 14(15):9680, 2022.

\bibitem{karim2022dynamic}
Muhammad~Monjurul Karim, Yu Li, Ruwen Qin, and Zhaozheng Yin.
\newblock A dynamic spatial-temporal attention network for early anticipation of traffic accidents.
\newblock {\em {IEEE} Trans. Intell. Transp. Syst.}, 23(7):9590--9600, 2022.

\bibitem{DBLP:journals/tits/KarimLQY22}
Muhammad~Monjurul Karim, Yu Li, Ruwen Qin, and Zhaozheng Yin.
\newblock A dynamic spatial-temporal attention network for early anticipation of traffic accidents.
\newblock {\em {IEEE} Trans. Intell. Transp. Syst.}, 23(7):9590--9600, 2022.

\bibitem{karim2022attention}
Muhammad~Monjurul Karim, Zhaozheng Yin, and Ruwen Qin.
\newblock An attention-guided multistream feature fusion network for early localization of risky traffic agents in driving videos.
\newblock {\em IEEE Trans. Intell. Veh. in Press}, 2023.

\bibitem{DBLP:conf/aaai/KimLHS19}
Hoon Kim, Kangwook Lee, Gyeongjo Hwang, and Changho Suh.
\newblock Crash to not crash: Learn to identify dangerous vehicles using a simulator.
\newblock In {\em AAAI}, pages 978--985, 2019.

\bibitem{kumeda2019vehicle}
Bulbula Kumeda et~al.
\newblock Vehicle accident and traffic classification using deep convolutional neural networks.
\newblock In {\em International Computer Conference on Wavelet Active Media Technology and Information Processing}, pages 323--328, 2019.

\bibitem{law2018cornernet}
Hei Law and Jia Deng.
\newblock Cornernet: Detecting objects as paired keypoints.
\newblock In {\em ECCV}, pages 765--781, 2018.

\bibitem{le2020hierarchical}
Thao~Minh Le, Vuong Le, Svetha Venkatesh, and Truyen Tran.
\newblock Hierarchical conditional relation networks for video question answering.
\newblock In {\em CVPR}, pages 9972--9981, 2020.

\bibitem{le2020attention}
Trung-Nghia Le, Shintaro Ono, Akihiro Sugimoto, and Hiroshi Kawasaki.
\newblock Attention {R-CNN} for accident detection.
\newblock In {\em IV}, pages 313--320, 2020.

\bibitem{lei2021less}
Jie Lei, Linjie Li, Luowei Zhou, Zhe Gan, Tamara~L Berg, Mohit Bansal, and Jingjing Liu.
\newblock Less is more: Clipbert for video-and-language learning via sparse sampling.
\newblock In {\em CVPR}, pages 7331--7341, 2021.

\bibitem{li2023gligen}
Yuheng Li, Haotian Liu, Qingyang Wu, Fangzhou Mu, Jianwei Yang, Jianfeng Gao, Chunyuan Li, and Yong~Jae Lee.
\newblock Gligen: Open-set grounded text-to-image generation.
\newblock In {\em Proceedings of the IEEE/CVF Conference on Computer Vision and Pattern Recognition}, pages 22511--22521, 2023.

\bibitem{lin2014microsoft}
Tsung-Yi Lin, Michael Maire, Serge Belongie, James Hays, Pietro Perona, Deva Ramanan, Piotr Doll{\'a}r, and C~Lawrence Zitnick.
\newblock Microsoft {COCO}: Common objects in context.
\newblock In {\em ECCV}, pages 740--755, 2014.

\bibitem{DBLP:journals/tits/LiuLCLX22}
Chunsheng Liu, Zijian Li, Faliang Chang, Shuang Li, and Jincan Xie.
\newblock Temporal shift and spatial attention-based two-stream network for traffic risk assessment.
\newblock {\em {IEEE} Trans. Intell. Transp. Syst.}, 23(8):12518--12530, 2022.

\bibitem{DBLP:journals/pami/LiuLL23}
Yang Liu, Guanbin Li, and Liang Lin.
\newblock Cross-modal causal relational reasoning for event-level visual question answering.
\newblock {\em {IEEE} Trans. Pattern Anal. Mach. Intell.}, 45(10):11624--11641, 2023.

\bibitem{liu2019roberta}
Yinhan Liu, Myle Ott, Naman Goyal, Jingfei Du, Mandar Joshi, Danqi Chen, Omer Levy, Mike Lewis, Luke Zettlemoyer, and Veselin Stoyanov.
\newblock Roberta: A robustly optimized bert pretraining approach.
\newblock {\em arXiv preprint arXiv:1907.11692}, 2019.

\bibitem{luoICASSP2023}
Haohan Luo and Feng Wang.
\newblock A simulation-based framework for urban traffic accident detection.
\newblock In {\em ICASSP}, pages 1--5, 2023.

\bibitem{DBLP:journals/iotj/MalawadeYHMKF22}
Arnav~Vaibhav Malawade, Shih{-}Yuan Yu, Brandon Hsu, Deepan Muthirayan, Pramod~P. Khargonekar, and Mohammad Abdullah~Al Faruque.
\newblock Spatiotemporal scene-graph embedding for autonomous vehicle collision prediction.
\newblock {\em {IEEE} Internet Things J.}, 9(12):9379--9388, 2022.

\bibitem{mildenhall2021nerf}
Ben Mildenhall, Pratul~P Srinivasan, Matthew Tancik, Jonathan~T Barron, Ravi Ramamoorthi, and Ren Ng.
\newblock Nerf: Representing scenes as neural radiance fields for view synthesis.
\newblock {\em Communications of the ACM}, 65(1):99--106, 2021.

\bibitem{ni2022expanding}
Bolin Ni, Houwen Peng, Minghao Chen, Songyang Zhang, Gaofeng Meng, Jianlong Fu, Shiming Xiang, and Haibin Ling.
\newblock Expanding language-image pretrained models for general video recognition.
\newblock In {\em European Conference on Computer Vision}, pages 1--18. Springer, 2022.

\bibitem{DBLP:journals/ict-express/PawarA22}
Karishma Pawar and Vahida Attar.
\newblock Deep learning based detection and localization of road accidents from traffic surveillance videos.
\newblock {\em {ICT} Express}, 8(3):379--387, 2022.

\bibitem{radford2021learning}
Alec Radford, Jong~Wook Kim, Chris Hallacy, Aditya Ramesh, Gabriel Goh, Sandhini Agarwal, Girish Sastry, Amanda Askell, Pamela Mishkin, Jack Clark, et~al.
\newblock Learning transferable visual models from natural language supervision.
\newblock In {\em ICML}, pages 8748--8763, 2021.

\bibitem{ren2015faster}
Shaoqing Ren, Kaiming He, Ross Girshick, and Jian Sun.
\newblock Faster {R-CNN}: Towards real-time object detection with region proposal networks.
\newblock {\em NeurIPS}, 28, 2015.

\bibitem{rombach2022high}
Robin Rombach, Andreas Blattmann, Dominik Lorenz, Patrick Esser, and Bj{\"o}rn Ommer.
\newblock High-resolution image synthesis with latent diffusion models.
\newblock In {\em CVPR}, pages 10684--10695, 2022.

\bibitem{roy2020detection}
Debaditya Roy, Tetsuhiro Ishizaka, C~Krishna Mohan, and Atsushi Fukuda.
\newblock Detection of collision-prone vehicle behavior at intersections using siamese interaction lstm.
\newblock {\em {IEEE} Trans. Intell. Transp. Syst.}, 23(4):3137--3147, 2020.

\bibitem{santhosh2021vehicular}
Kelathodi~Kumaran Santhosh, Debi~Prosad Dogra, Partha~Pratim Roy, and Adway Mitra.
\newblock Vehicular trajectory classification and traffic anomaly detection in videos using a hybrid cnn-vae architecture.
\newblock {\em {IEEE} Trans. Intell. Transp. Syst.}, 23(8):11891--11902, 2021.

\bibitem{DBLP:journals/tits/SinghM19}
Dinesh Singh and Chalavadi~Krishna Mohan.
\newblock Deep spatio-temporal representation for detection of road accidents using stacked autoencoder.
\newblock {\em {IEEE} Trans. Intell. Transp. Syst.}, 20(3):879--887, 2019.

\bibitem{DBLP:conf/iclr/SongME21}
Jiaming Song, Chenlin Meng, and Stefano Ermon.
\newblock Denoising diffusion implicit models.
\newblock In {\em ICLR}, 2021.

\bibitem{suzuki2018anticipating}
Tomoyuki Suzuki, Hirokatsu Kataoka, Yoshimitsu Aoki, and Yutaka Satoh.
\newblock Anticipating traffic accidents with adaptive loss and large-scale incident {DB}.
\newblock In {\em CVPR}, pages 3521--3529, 2018.

\bibitem{taccari2018classification}
Leonardo Taccari, Francesco Sambo, Luca Bravi, Samuele Salti, Leonardo Sarti, Matteo Simoncini, and Alessandro Lori.
\newblock Classification of crash and near-crash events from dashcam videos and telematics.
\newblock In {\em ITSC}, pages 2460--2465, 2018.

\bibitem{DBLP:conf/icml/TanL19}
Mingxing Tan and Quoc~V. Le.
\newblock Efficientnet: Rethinking model scaling for convolutional neural networks.
\newblock In Kamalika Chaudhuri and Ruslan Salakhutdinov, editors, {\em ICML}, volume~97, pages 6105--6114, 2019.

\bibitem{ultralyticsgithub}
Ultralytics.
\newblock Ultralytics github repository.
\newblock \url{https://github.com/ultralytics/ultralytics}, November 2023.

\bibitem{unterthiner2018towards}
Thomas Unterthiner, Sjoerd Van~Steenkiste, Karol Kurach, Raphael Marinier, Marcin Michalski, and Sylvain Gelly.
\newblock Towards accurate generative models of video: A new metric \& challenges.
\newblock {\em arXiv preprint arXiv:1812.01717}, 2018.

\bibitem{DBLP:journals/tits/VijayDCNK23}
Thakare~Kamalakar Vijay, Debi~Prosad Dogra, Heeseung Choi, Gi~Pyo Nam, and Ig{-}Jae Kim.
\newblock Detection of road accidents using synthetically generated multi-perspective accident videos.
\newblock {\em {IEEE} Trans. Intell. Transp. Syst.}, 24(2):1926--1935, 2023.

\bibitem{voleti2022mcvd}
Vikram Voleti, Alexia Jolicoeur-Martineau, and Chris Pal.
\newblock Mcvd-masked conditional video diffusion for prediction, generation, and interpolation.
\newblock {\em NeurIPS}, 35:23371--23385, 2022.

\bibitem{vu2019cascade}
Thang Vu, Hyunjun Jang, Trung~X. Pham, and Chang~Dong Yoo.
\newblock Cascade {RPN:} delving into high-quality region proposal network with adaptive convolution.
\newblock In {\em NeurIPS}, pages 1430--1440, 2019.

\bibitem{wang2023gsc}
Tianhang Wang, Kai Chen, Guang Chen, Bin Li, Zhijun Li, Zhengfa Liu, and Changjun Jiang.
\newblock {GSC}: A graph and spatio-temporal continuity based framework for accident anticipation.
\newblock {\em {IEEE} Trans. Intell. Veh. in Press}, 2023.

\bibitem{DBLP:journals/corr/abs-2304-01168}
Tianqi Wang, Sukmin Kim, Wenxuan Ji, Enze Xie, Chongjian Ge, Junsong Chen, Zhenguo Li, and Ping Luo.
\newblock Deepaccident: {A} motion and accident prediction benchmark for {V2X} autonomous driving.
\newblock {\em CoRR}, abs/2304.01168, 2023.

\bibitem{wu2023tune}
Jay~Zhangjie Wu, Yixiao Ge, Xintao Wang, Stan~Weixian Lei, Yuchao Gu, Yufei Shi, Wynne Hsu, Ying Shan, Xiaohu Qie, and Mike~Zheng Shou.
\newblock Tune-a-{V}ideo: One-shot tuning of image diffusion models for text-to-video generation.
\newblock In {\em ICCV}, pages 7623--7633, 2023.

\bibitem{xiao2023can}
Junbin Xiao, Angela Yao, Yicong Li, and Tat~Seng Chua.
\newblock Can i trust your answer? visually grounded video question answering.
\newblock {\em arXiv preprint arXiv:2309.01327}, 2023.

\bibitem{xiao2022video}
Junbin Xiao, Pan Zhou, Tat-Seng Chua, and Shuicheng Yan.
\newblock Video graph transformer for video question answering.
\newblock In {\em ECCV}, pages 39--58, 2022.

\bibitem{DBLP:journals/pami/XiaoZYLHYC23}
Junbin Xiao, Pan Zhou, Angela Yao, Yicong Li, Richang Hong, Shuicheng Yan, and Tat{-}Seng Chua.
\newblock Contrastive video question answering via video graph transformer.
\newblock {\em {IEEE} T-PAMI}, 45(11):13265--13280, 2023.

\bibitem{DBLP:conf/cvpr/XuHL21}
Li Xu, He Huang, and Jun Liu.
\newblock {SUTD}-{T}raffic{QA}: {A} question answering benchmark and an efficient network for video reasoning over traffic events.
\newblock In {\em CVPR}, pages 9878--9888, 2021.

\bibitem{yao2022dota}
Yu Yao, Xizi Wang, Mingze Xu, Zelin Pu, Yuchen Wang, Ella~M. Atkins, and David~J. Crandall.
\newblock Dota: Unsupervised detection of traffic anomaly in driving videos.
\newblock {\em {IEEE} Trans. Pattern Anal. Mach. Intell.}, 45(1):444--459, 2023.

\bibitem{DBLP:conf/iros/YaoXWCA19}
Yu Yao, Mingze Xu, Yuchen Wang, David~J. Crandall, and Ella~M. Atkins.
\newblock Unsupervised traffic accident detection in first-person videos.
\newblock In {\em IROS}, pages 273--280, 2019.

\bibitem{DBLP:conf/eccv/YouH20}
Tackgeun You and Bohyung Han.
\newblock Traffic accident benchmark for causality recognition.
\newblock In {\em ECCV}, volume 12352, pages 540--556, 2020.

\bibitem{DBLP:conf/cvpr/YuCWXCLMD20}
Fisher Yu, Haofeng Chen, Xin Wang, Wenqi Xian, Yingying Chen, Fangchen Liu, Vashisht Madhavan, and Trevor Darrell.
\newblock {BDD100K:} {A} diverse driving dataset for heterogeneous multitask learning.
\newblock In {\em CVPR}, pages 2633--2642, 2020.

\bibitem{yu2023self}
Shoubin Yu, Jaemin Cho, Prateek Yadav, and Mohit Bansal.
\newblock Self-chained image-language model for video localization and question answering.
\newblock {\em NeurIPS}, 2023.

\bibitem{DBLP:conf/cvpr/ZangWPL23}
Chuanqi Zang, Hanqing Wang, Mingtao Pei, and Wei Liang.
\newblock Discovering the real association: Multimodal causal reasoning in video question answering.
\newblock In {\em CVPR}, pages 19027--19036, 2023.

\bibitem{zhang2023controlvideo}
Yabo Zhang, Yuxiang Wei, Dongsheng Jiang, Xiaopeng Zhang, Wangmeng Zuo, and Qi Tian.
\newblock Controlvideo: Training-free controllable text-to-video generation.
\newblock {\em arXiv preprint arXiv:2305.13077}, 2023.

\bibitem{zhou2022spatio}
Zhili Zhou, Xiaohua Dong, Zhetao Li, Keping Yu, Chun Ding, and Yimin Yang.
\newblock Spatio-temporal feature encoding for traffic accident detection in vanet environment.
\newblock {\em {IEEE} Trans. Intell. Transp. Syst.}, 23(10):19772--19781, 2022.

\bibitem{zhu2021deformable}
Xizhou Zhu, Weijie Su, Lewei Lu, Bin Li, Xiaogang Wang, and Jifeng Dai.
\newblock Deformable detr: Deformable transformers for end-to-end object detection.
\newblock In {\em ICLR}, 2021.

\end{thebibliography}
}
\clearpage
\setcounter{equation}{0}
\setcounter{figure}{0}
\setcounter{table}{0}
\setcounter{page}{1}
\setcounter{section}{0}

\twocolumn[{
\renewcommand\twocolumn[1][]{#1}
\maketitle
\begin{center}
    \textbf{\Large Supplementary Material of Abductive Ego-View Accident Video Understanding for Safe Driving Perception}
    \vspace{20pt}
     \centering
    \captionsetup{type=figure}
\includegraphics[width=\linewidth]{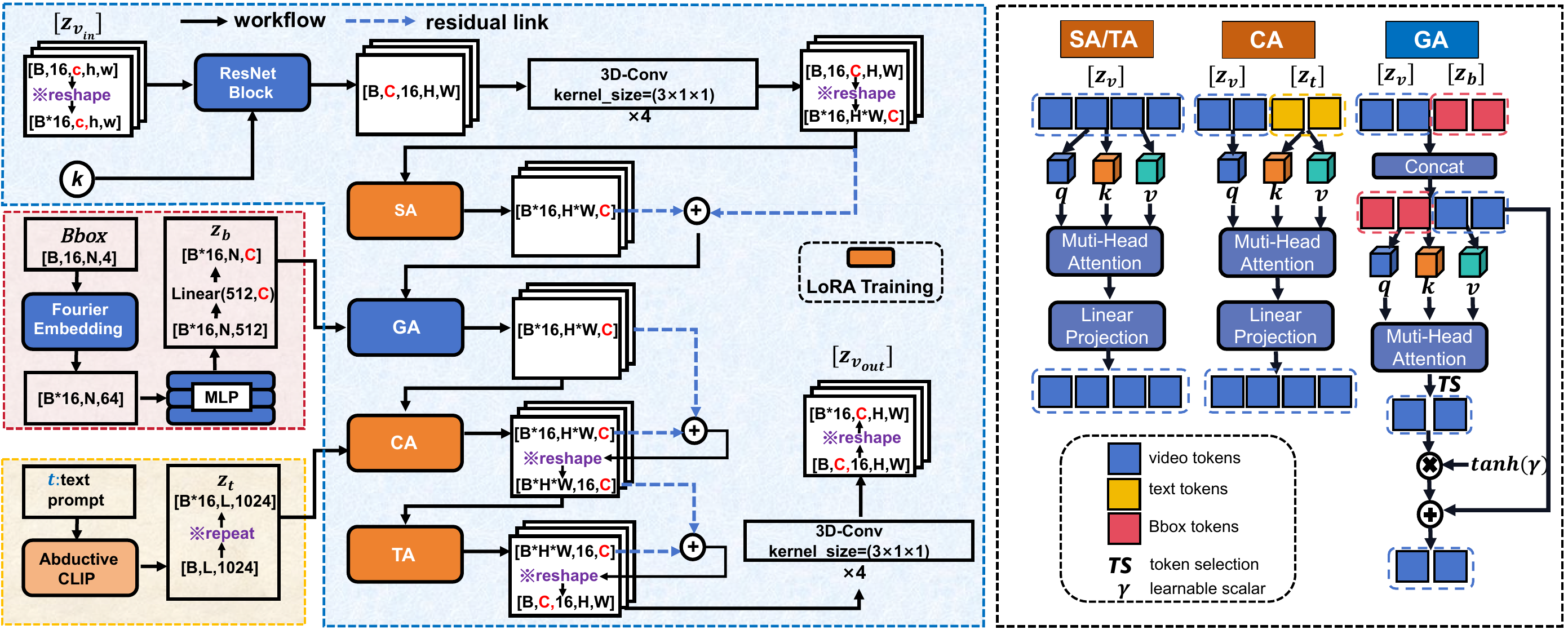}
   \caption{\small{\textbf{The detailed workflow of 3D-CAB} in OAVD. 3D-CAB is one layer of the 3D U-net $\phi$ in Fig. \ref{fig5} of the main paper body. To be clear, we denote the input representation of 3D-CAB as $\textbf{z}_{v_{in}}$}, and the output representation as $\textbf{z}_{v_{out}}$. Within 3D-CAB, the feature representation of bounding boxes $\textbf{z}_b$ and text descriptions $\textbf{z}_t$ are fused successively to the Gated Self-Attention (GA) and Cross-Attention (CA) modules, where $\textbf{z}_b$ is obtained by \underline{$\text{MLP}(\text{Fourier}(Bbox))$} in Eq.(4) and $\textbf{z}_t$ is generated by our Abductive CLIP. In \underline{\text{Fourier}(Bbox)}, there is a Token Selection (TS) module \cite{li2023gligen} to find the important tokens for object representation learning. Notably, different from \cite{wu2023tune}, the query ($\textbf{q}$), key ($\textbf{k}$), and value ($\textbf{v}$) are all updated in the OAVD training phase.}
   \label{fig9}
\end{center}}]
\maketitle

 \begin{table*}[!t]\footnotesize
\centering
\caption{The object detection results of V1-Train [\YellowRect, \BlueRect, \PurpleRect]) and V2-Train [\YellowRect, \BlueRect]) for 11 state-of-the-art detectors on the \datasetname, \emph{w.r.t.}, \textbf{pedestrians}, \textbf{cars}, \textbf{motorcycles}, and \textbf{trucks}. }
\begin{tabular}{c|cccccc||cccccc}
\multicolumn{13}{c}{\textbf{pedestrian}}                                                                                                                           \\     \toprule
\multicolumn{1}{c}{}                  & \multicolumn{6}{|c|}{V1-Train [\YellowRect, \BlueRect, \PurpleRect], test. [\YellowRect, \BlueRect, \PurpleRect]}                                                                                                                                        & \multicolumn{6}{c}{V2-Train [\YellowRect, \BlueRect], test. [\PurpleRect]}                                                                                                                     \\ \hline
\multicolumn{1}{c|}{method}            & \multicolumn{1}{c|}{AP}   & \multicolumn{1}{c|}{AP50} & \multicolumn{1}{c|}{mAP75} & \multicolumn{1}{c|}{AP\_S} & \multicolumn{1}{c|}{AP\_M} & \multicolumn{1}{c|}{AP\_L} & \multicolumn{1}{c|}{AP}   & \multicolumn{1}{c|}{AP50} & \multicolumn{1}{c|}{AP75} & \multicolumn{1}{c|}{AP\_S}  & \multicolumn{1}{c|}{AP\_M} & AP\_L  \\ \hline
\multicolumn{1}{c|}{FasterRCNN \cite{ren2015faster} }      & \multicolumn{1}{c|}{0.454} & \multicolumn{1}{c|}{0.715} & \multicolumn{1}{c|}{0.522} & \multicolumn{1}{c|}{0.491} & \multicolumn{1}{c|}{0.473} & \multicolumn{1}{c|}{0.376} & \multicolumn{1}{c|}{0.294} & \multicolumn{1}{c|}{0.535} & \multicolumn{1}{c|}{0.302} & \multicolumn{1}{c|}{0.33}   & \multicolumn{1}{c|}{0.340}  & 0.149  \\ \hline
\multicolumn{1}{c|}{CornerNet \cite{law2018cornernet}}         & \multicolumn{1}{c|}{0.378} & \multicolumn{1}{c|}{0.549} & \multicolumn{1}{c|}{0.439} & \multicolumn{1}{c|}{0.317} & \multicolumn{1}{c|}{0.436} & \multicolumn{1}{c|}{0.252} & \multicolumn{1}{c|}{0.335} & \multicolumn{1}{c|}{0.511} & \multicolumn{1}{c|}{0.384} & \multicolumn{1}{c|}{0.203}  & \multicolumn{1}{c|}{0.406} & 0.217  \\ \hline
\multicolumn{1}{c|}{CascadeRPN \cite{vu2019cascade}}        & \multicolumn{1}{c|}{0.448} & \multicolumn{1}{c|}{0.699} & \multicolumn{1}{c|}{0.513} & \multicolumn{1}{c|}{0.443} & \multicolumn{1}{c|}{0.46}  & \multicolumn{1}{c|}{0.424} & \multicolumn{1}{c|}{0.365} & \multicolumn{1}{c|}{0.593} & \multicolumn{1}{c|}{0.407} & \multicolumn{1}{c|}{0.331}  & \multicolumn{1}{c|}{0.405} & 0.266  \\ \hline
\multicolumn{1}{c|}{CenterNet \cite{duan2019centernet} }         & \multicolumn{1}{c|}{0.011} & \multicolumn{1}{c|}{0.040}  & \multicolumn{1}{c|}{0.002} & \multicolumn{1}{c|}{0.037} & \multicolumn{1}{c|}{0.014} & \multicolumn{1}{c|}{0.011} & \multicolumn{1}{c|}{0.047} & \multicolumn{1}{c|}{0.135} & \multicolumn{1}{c|}{0.019} & \multicolumn{1}{c|}{0.034}  & \multicolumn{1}{c|}{0.062} & 0.021  \\ \hline
\multicolumn{1}{c|}{DETR \cite{detr2020}}              & \multicolumn{1}{c|}{0.099} & \multicolumn{1}{c|}{0.294} & \multicolumn{1}{c|}{0.038} & \multicolumn{1}{c|}{0.096} & \multicolumn{1}{c|}{0.088} & \multicolumn{1}{c|}{0.15}  & \multicolumn{1}{c|}{0.058} & \multicolumn{1}{c|}{0.175} & \multicolumn{1}{c|}{0.022} & \multicolumn{1}{c|}{0.029}  & \multicolumn{1}{c|}{0.064} & 0.067  \\ \hline
\multicolumn{1}{c|}{EfficientNet \cite{DBLP:conf/icml/TanL19}}      & \multicolumn{1}{c|}{0.114} & \multicolumn{1}{c|}{0.299} & \multicolumn{1}{c|}{0.055} & \multicolumn{1}{c|}{0.096} & \multicolumn{1}{c|}{0.127} & \multicolumn{1}{c|}{0.106} & \multicolumn{1}{c|}{0.000}     & \multicolumn{1}{c|}{0.002} & \multicolumn{1}{c|}{0.000}     & \multicolumn{1}{c|}{0.004}  & \multicolumn{1}{c|}{0.000}     & 0.001  \\ \hline
\multicolumn{1}{c|}{Deformable-DeTR \cite{zhu2021deformable}} & \multicolumn{1}{c|}{0.404} & \multicolumn{1}{c|}{0.686} & \multicolumn{1}{c|}{0.441} & \multicolumn{1}{c|}{0.396} & \multicolumn{1}{c|}{0.421} & \multicolumn{1}{c|}{0.361} & \multicolumn{1}{c|}{0.369} & \multicolumn{1}{c|}{0.64}  & \multicolumn{1}{c|}{0.404} & \multicolumn{1}{c|}{0.317}  & \multicolumn{1}{c|}{0.414} & 0.279  \\ \hline
\multicolumn{1}{c|}{YOLOx \cite{DBLP:journals/corr/abs-2107-08430}}             & \multicolumn{1}{c|}{0.424} & \multicolumn{1}{c|}{0.695} & \multicolumn{1}{c|}{0.471} & \multicolumn{1}{c|}{0.387} & \multicolumn{1}{c|}{0.440}  & \multicolumn{1}{c|}{0.406} & \multicolumn{1}{c|}{0.293} & \multicolumn{1}{c|}{0.531} & \multicolumn{1}{c|}{0.297} & \multicolumn{1}{c|}{0.206}  & \multicolumn{1}{c|}{0.344} & 0.213  \\ \hline
\multicolumn{1}{c|}{YOLOv5s \cite{glenn_jocher_2022_7002879} }           & \multicolumn{1}{c|}{\textbf{0.529}} & \multicolumn{1}{c|}{\textbf{0.784}} & \multicolumn{1}{c|}{\textbf{0.632}} & \multicolumn{1}{c|}{0.459} & \multicolumn{1}{c|}{\textbf{0.544}} & \multicolumn{1}{c|}{\textbf{0.521}} & \multicolumn{1}{c|}{0.370}  & \multicolumn{1}{c|}{0.632} & \multicolumn{1}{c|}{0.412} & \multicolumn{1}{c|}{0.295}  & \multicolumn{1}{c|}{0.419} & 0.265  \\ \hline
\multicolumn{1}{c|}{DiffusionDet  \cite{chen2023diffusiondet} }      & \multicolumn{1}{c|}{0.527} & \multicolumn{1}{c|}{0.767} & \multicolumn{1}{c|}{0.607} & \multicolumn{1}{c|}{\textbf{0.463}} & \multicolumn{1}{c|}{\textbf{0.544}} & \multicolumn{1}{c|}{0.516} & \multicolumn{1}{c|}{\textbf{0.480}}  & \multicolumn{1}{c|}{\textbf{0.699}} & \multicolumn{1}{c|}{\textbf{0.557}} & \multicolumn{1}{c|}{\textbf{0.423}}  & \multicolumn{1}{c|}{\textbf{0.531}} & \textbf{0.376}  \\ \hline
\multicolumn{1}{c|}{YOLOv8 \cite{ultralyticsgithub}}            & \multicolumn{1}{c|}{0.506} & \multicolumn{1}{c|}{0.748} & \multicolumn{1}{c|}{0.590}  & \multicolumn{1}{c|}{0.455} & \multicolumn{1}{c|}{0.516} & \multicolumn{1}{c|}{0.512} & \multicolumn{1}{c|}{0.415} & \multicolumn{1}{c|}{0.650}  & \multicolumn{1}{c|}{0.481} & \multicolumn{1}{c|}{0.322}  & \multicolumn{1}{c|}{0.463} & 0.337  \\ 
\toprule
\multicolumn{13}{c}{\textbf{car}} \\     \toprule
\multicolumn{1}{c}{}                  & \multicolumn{6}{|c|}{V1-Train [\YellowRect, \BlueRect, \PurpleRect], test. [\YellowRect, \BlueRect, \PurpleRect]}                                                                                                                                        & \multicolumn{6}{c}{V2-Train [\YellowRect, \BlueRect], test. [\PurpleRect]}                                                                                                                     \\ \hline
\multicolumn{1}{c|}{Detectors}            & \multicolumn{1}{c|}{AP}   & \multicolumn{1}{c|}{AP50} & \multicolumn{1}{c|}{AP75} & \multicolumn{1}{c|}{AP\_S} & \multicolumn{1}{c|}{AP\_M} & \multicolumn{1}{c|}{AP\_L} & \multicolumn{1}{c|}{AP}   & \multicolumn{1}{c|}{AP50} & \multicolumn{1}{c|}{AP75} & \multicolumn{1}{c|}{AP\_S}  & \multicolumn{1}{c|}{AP\_M} & AP\_L  \\ \hline
\multicolumn{1}{c|}{FasterRCNN \cite{ren2015faster} }      & \multicolumn{1}{c|}{0.677} & \multicolumn{1}{c|}{0.910}  & \multicolumn{1}{c|}{0.788} & \multicolumn{1}{c|}{0.532} & \multicolumn{1}{c|}{0.672} & \multicolumn{1}{c|}{0.771} & \multicolumn{1}{c|}{0.608} & \multicolumn{1}{c|}{0.851} & \multicolumn{1}{c|}{0.694} & \multicolumn{1}{c|}{0.501}  & \multicolumn{1}{c|}{0.629} & 0.639  \\ \hline
\multicolumn{1}{c|}{CornerNet \cite{law2018cornernet}}         & \multicolumn{1}{c|}{0.493} & \multicolumn{1}{c|}{0.628} & \multicolumn{1}{c|}{0.537} & \multicolumn{1}{c|}{0.259} & \multicolumn{1}{c|}{0.561} & \multicolumn{1}{c|}{0.532} & \multicolumn{1}{c|}{0.481} & \multicolumn{1}{c|}{0.639} & \multicolumn{1}{c|}{0.522} & \multicolumn{1}{c|}{0.259}  & \multicolumn{1}{c|}{0.563} & 0.479  \\ \hline
\multicolumn{1}{c|}{CascadeRPN \cite{vu2019cascade} }        & \multicolumn{1}{c|}{0.714} & \multicolumn{1}{c|}{0.908} & \multicolumn{1}{c|}{0.805} & \multicolumn{1}{c|}{0.567} & \multicolumn{1}{c|}{0.701} & \multicolumn{1}{c|}{0.819} & \multicolumn{1}{c|}{0.644} & \multicolumn{1}{c|}{0.866} & \multicolumn{1}{c|}{0.733} & \multicolumn{1}{c|}{0.531}  & \multicolumn{1}{c|}{0.646} & 0.706  \\ \hline
\multicolumn{1}{c|}{CenterNet \cite{duan2019centernet} }         & \multicolumn{1}{c|}{0.073} & \multicolumn{1}{c|}{0.135} & \multicolumn{1}{c|}{0.071} & \multicolumn{1}{c|}{0.100}   & \multicolumn{1}{c|}{0.094} & \multicolumn{1}{c|}{0.062} & \multicolumn{1}{c|}{0.264} & \multicolumn{1}{c|}{0.515} & \multicolumn{1}{c|}{0.242} & \multicolumn{1}{c|}{0.194}  & \multicolumn{1}{c|}{0.328} & 0.256  \\ \hline
\multicolumn{1}{c|}{DETR \cite{detr2020}}              & \multicolumn{1}{c|}{0.402} & \multicolumn{1}{c|}{0.746} & \multicolumn{1}{c|}{0.381} & \multicolumn{1}{c|}{0.133} & \multicolumn{1}{c|}{0.349} & \multicolumn{1}{c|}{0.638} & \multicolumn{1}{c|}{0.346} & \multicolumn{1}{c|}{0.676} & \multicolumn{1}{c|}{0.312} & \multicolumn{1}{c|}{0.135}  & \multicolumn{1}{c|}{0.308} & 0.524  \\ \hline
\multicolumn{1}{c|}{EfficientNet \cite{DBLP:conf/icml/TanL19}}      & \multicolumn{1}{c|}{0.409} & \multicolumn{1}{c|}{0.745} & \multicolumn{1}{c|}{0.426} & \multicolumn{1}{c|}{0.140}  & \multicolumn{1}{c|}{0.423} & \multicolumn{1}{c|}{0.547} & \multicolumn{1}{c|}{0.146} & \multicolumn{1}{c|}{0.359} & \multicolumn{1}{c|}{0.086} & \multicolumn{1}{c|}{0.050}   & \multicolumn{1}{c|}{0.151} & 0.191  \\ \hline
\multicolumn{1}{c|}{Deformable-DeTR \cite{zhu2021deformable}} & \multicolumn{1}{c|}{0.657} & \multicolumn{1}{c|}{0.906} & \multicolumn{1}{c|}{0.763} & \multicolumn{1}{c|}{0.466} & \multicolumn{1}{c|}{0.636} & \multicolumn{1}{c|}{0.801} & \multicolumn{1}{c|}{0.607} & \multicolumn{1}{c|}{0.882} & \multicolumn{1}{c|}{0.684} & \multicolumn{1}{c|}{0.393}  & \multicolumn{1}{c|}{0.599} & 0.736  \\ \hline
\multicolumn{1}{c|}{YOLOx \cite{DBLP:journals/corr/abs-2107-08430}}             & \multicolumn{1}{c|}{0.713} & \multicolumn{1}{c|}{0.913} & \multicolumn{1}{c|}{0.799} & \multicolumn{1}{c|}{0.529} & \multicolumn{1}{c|}{0.706} & \multicolumn{1}{c|}{0.840}  & \multicolumn{1}{c|}{0.619} & \multicolumn{1}{c|}{0.844} & \multicolumn{1}{c|}{0.692} & \multicolumn{1}{c|}{0.431}  & \multicolumn{1}{c|}{0.622} & 0.720   \\ \hline
\multicolumn{1}{c|}{YOLOv5s \cite{glenn_jocher_2022_7002879} }           & \multicolumn{1}{c|}{\textbf{0.769}} & \multicolumn{1}{c|}{\textbf{0.936}} & \multicolumn{1}{c|}{\textbf{0.862}} & \multicolumn{1}{c|}{0.585} & \multicolumn{1}{c|}{\textbf{0.762}} & \multicolumn{1}{c|}{\textbf{0.882}} & \multicolumn{1}{c|}{0.682} & \multicolumn{1}{c|}{0.902} & \multicolumn{1}{c|}{0.787} & \multicolumn{1}{c|}{0.495}  & \multicolumn{1}{c|}{0.684} & 0.773  \\ \hline
\multicolumn{1}{c|}{DiffusionDet  \cite{chen2023diffusiondet} }      & \multicolumn{1}{c|}{0.754} & \multicolumn{1}{c|}{0.932} & \multicolumn{1}{c|}{0.836} & \multicolumn{1}{c|}{\textbf{0.586}} & \multicolumn{1}{c|}{0.747} & \multicolumn{1}{c|}{0.867} & \multicolumn{1}{c|}{\textbf{0.720}}  & \multicolumn{1}{c|}{\textbf{0.908}} & \multicolumn{1}{c|}{\textbf{0.801}} & \multicolumn{1}{c|}{\textbf{0.575}}  & \multicolumn{1}{c|}{\textbf{0.721}} & \textbf{0.808}  \\ \hline
\multicolumn{1}{c|}{YOLOv8 \cite{ultralyticsgithub}}            & \multicolumn{1}{c|}{0.755} & \multicolumn{1}{c|}{0.926} & \multicolumn{1}{c|}{0.836} & \multicolumn{1}{c|}{0.576} & \multicolumn{1}{c|}{0.748} & \multicolumn{1}{c|}{0.867} & \multicolumn{1}{c|}{0.707} & \multicolumn{1}{c|}{0.896} & \multicolumn{1}{c|}{0.791} & \multicolumn{1}{c|}{0.532}  & \multicolumn{1}{c|}{0.706} & 0.801  \\ 
\toprule
\multicolumn{13}{c}{\textbf{motorcycle}}                                                                                                                                                                  \\     \toprule
\multicolumn{1}{c}{}                  & \multicolumn{6}{|c|}{V1-Train [\YellowRect, \BlueRect, \PurpleRect], test. [\YellowRect, \BlueRect, \PurpleRect]}                                                                                                                                        & \multicolumn{6}{c}{V2-Train [\YellowRect, \BlueRect], test. [\PurpleRect]}                                                                                                                     \\ \hline
\multicolumn{1}{c|}{Detectors}   & \multicolumn{1}{c|}{AP}   & \multicolumn{1}{c|}{AP50} & \multicolumn{1}{c|}{AP75} & \multicolumn{1}{c|}{AP\_S} & \multicolumn{1}{c|}{AP\_M} & \multicolumn{1}{c|}{AP\_L} & \multicolumn{1}{c|}{AP}   & \multicolumn{1}{c|}{AP50} & \multicolumn{1}{c|}{AP75} & \multicolumn{1}{c|}{AP\_S}  & \multicolumn{1}{c|}{AP\_M} & AP\_L  \\ \hline
\multicolumn{1}{c|}{FasterRCNN \cite{ren2015faster} }      & \multicolumn{1}{c|}{0.316} & \multicolumn{1}{c|}{0.554} & \multicolumn{1}{c|}{0.330}  & \multicolumn{1}{c|}{0.268} & \multicolumn{1}{c|}{0.341} & \multicolumn{1}{c|}{0.291} & \multicolumn{1}{c|}{0.165} & \multicolumn{1}{c|}{0.342} & \multicolumn{1}{c|}{0.139} & \multicolumn{1}{c|}{0.208}  & \multicolumn{1}{c|}{0.200}   & 0.081  \\ \hline
\multicolumn{1}{c|}{CornerNet \cite{law2018cornernet}}         & \multicolumn{1}{c|}{0.232} & \multicolumn{1}{c|}{0.393} & \multicolumn{1}{c|}{0.250}  & \multicolumn{1}{c|}{0.200}   & \multicolumn{1}{c|}{0.284} & \multicolumn{1}{c|}{0.147} & \multicolumn{1}{c|}{0.176} & \multicolumn{1}{c|}{0.334} & \multicolumn{1}{c|}{0.175} & \multicolumn{1}{c|}{0.160}   & \multicolumn{1}{c|}{0.222} & 0.108  \\ \hline
\multicolumn{1}{c|}{CascadeRPN \cite{vu2019cascade} }        & \multicolumn{1}{c|}{0.320}  & \multicolumn{1}{c|}{0.511} & \multicolumn{1}{c|}{0.340}  & \multicolumn{1}{c|}{0.272} & \multicolumn{1}{c|}{0.336} & \multicolumn{1}{c|}{0.313} & \multicolumn{1}{c|}{0.175} & \multicolumn{1}{c|}{0.357} & \multicolumn{1}{c|}{0.150}  & \multicolumn{1}{c|}{0.186}  & \multicolumn{1}{c|}{0.200}   & 0.0153 \\ \hline
\multicolumn{1}{c|}{CenterNet \cite{duan2019centernet} }         & \multicolumn{1}{c|}{0.002} & \multicolumn{1}{c|}{0.008} & \multicolumn{1}{c|}{0.001} & \multicolumn{1}{c|}{0.021} & \multicolumn{1}{c|}{0.003} & \multicolumn{1}{c|}{0.001} & \multicolumn{1}{c|}{0.016} & \multicolumn{1}{c|}{0.052} & \multicolumn{1}{c|}{0.005} & \multicolumn{1}{c|}{0.053}  & \multicolumn{1}{c|}{0.019} & 0.005  \\ \hline
\multicolumn{1}{c|}{DETR \cite{detr2020}}              & \multicolumn{1}{c|}{0.115} & \multicolumn{1}{c|}{0.306} & \multicolumn{1}{c|}{0.059} & \multicolumn{1}{c|}{0.057} & \multicolumn{1}{c|}{0.123} & \multicolumn{1}{c|}{0.128} & \multicolumn{1}{c|}{0.038} & \multicolumn{1}{c|}{0.121} & \multicolumn{1}{c|}{0.010}  & \multicolumn{1}{c|}{0.029}  & \multicolumn{1}{c|}{0.044} & 0.035  \\ \hline
\multicolumn{1}{c|}{EfficientNet \cite{DBLP:conf/icml/TanL19}}      & \multicolumn{1}{c|}{0.133} & \multicolumn{1}{c|}{0.312} & \multicolumn{1}{c|}{0.085} & \multicolumn{1}{c|}{0.074} & \multicolumn{1}{c|}{0.151} & \multicolumn{1}{c|}{0.127} & \multicolumn{1}{c|}{0.002} & \multicolumn{1}{c|}{0.006} & \multicolumn{1}{c|}{0.000}     & \multicolumn{1}{c|}{0.014}  & \multicolumn{1}{c|}{0.002} & 0.001  \\ \hline
\multicolumn{1}{c|}{Deformable-DeTR \cite{zhu2021deformable}} & \multicolumn{1}{c|}{0.276} & \multicolumn{1}{c|}{0.506} & \multicolumn{1}{c|}{0.276} & \multicolumn{1}{c|}{0.231} & \multicolumn{1}{c|}{0.305} & \multicolumn{1}{c|}{0.266} & \multicolumn{1}{c|}{0.201} & \multicolumn{1}{c|}{0.417} & \multicolumn{1}{c|}{0.173} & \multicolumn{1}{c|}{0.115} & \multicolumn{1}{c|}{0.223} & 0.176  \\ \hline
\multicolumn{1}{c|}{YOLOx \cite{DBLP:journals/corr/abs-2107-08430}}             & \multicolumn{1}{c|}{0.332} & \multicolumn{1}{c|}{0.560}  & \multicolumn{1}{c|}{0.356} & \multicolumn{1}{c|}{0.253} & \multicolumn{1}{c|}{0.365} & \multicolumn{1}{c|}{0.312} & \multicolumn{1}{c|}{0.148} & \multicolumn{1}{c|}{0.318} & \multicolumn{1}{c|}{0.120}  & \multicolumn{1}{c|}{0.183}  & \multicolumn{1}{c|}{0.189} & 0.125  \\ \hline
\multicolumn{1}{c|}{YOLOv5s \cite{glenn_jocher_2022_7002879} }           & \multicolumn{1}{c|}{\textbf{0.388}} & \multicolumn{1}{c|}{\textbf{0.615}} & \multicolumn{1}{c|}{\textbf{0.429}} & \multicolumn{1}{c|}{\textbf{0.301}} & \multicolumn{1}{c|}{\textbf{0.406}} & \multicolumn{1}{c|}{\textbf{0.391}} & \multicolumn{1}{c|}{0.061} & \multicolumn{1}{c|}{0.146} & \multicolumn{1}{c|}{0.040}  & \multicolumn{1}{c|}{0.017}  & \multicolumn{1}{c|}{0.044} & 0.105  \\ \hline
\multicolumn{1}{c|}{DiffusionDet  \cite{chen2023diffusiondet} }      & \multicolumn{1}{c|}{0.375} & \multicolumn{1}{c|}{0.599} & \multicolumn{1}{c|}{0.403} & \multicolumn{1}{c|}{0.300}   & \multicolumn{1}{c|}{0.398} & \multicolumn{1}{c|}{0.365} & \multicolumn{1}{c|}{\textbf{0.286}} & \multicolumn{1}{c|}{\textbf{0.493}} & \multicolumn{1}{c|}{\textbf{0.297}} & \multicolumn{1}{c|}{\textbf{0.256}}  & \multicolumn{1}{c|}{\textbf{0.325}} & \textbf{0.219}  \\ \hline
\multicolumn{1}{c|}{YOLOv8 \cite{ultralyticsgithub}}            & \multicolumn{1}{c|}{0.370}  & \multicolumn{1}{c|}{0.578} & \multicolumn{1}{c|}{0.412} & \multicolumn{1}{c|}{0.296} & \multicolumn{1}{c|}{0.390}  & \multicolumn{1}{c|}{0.368} & \multicolumn{1}{c|}{0.241} & \multicolumn{1}{c|}{0.440}  & \multicolumn{1}{c|}{0.237} & \multicolumn{1}{c|}{0.241}  & \multicolumn{1}{c|}{0.271} & 0.215  \\ 
\toprule
\multicolumn{13}{c}{\textbf{truck}} \\     \toprule
\multicolumn{1}{c}{}                  & \multicolumn{6}{|c|}{V1-Train [\YellowRect, \BlueRect, \PurpleRect], test. [\YellowRect, \BlueRect, \PurpleRect]}                                                                                                                                        & \multicolumn{6}{c}{V2-Train [\YellowRect, \BlueRect], test. [\PurpleRect]}                                                                                                                     \\ \hline
\multicolumn{1}{c}{Detectors}   & \multicolumn{1}{c|}{AP}   & \multicolumn{1}{c|}{AP50} & \multicolumn{1}{c|}{AP75} & \multicolumn{1}{c|}{AP\_S} & \multicolumn{1}{c|}{AP\_M} & \multicolumn{1}{c|}{AP\_L} & \multicolumn{1}{c|}{AP}   & \multicolumn{1}{c|}{AP50} & \multicolumn{1}{c|}{AP75} & \multicolumn{1}{c|}{AP\_S}  & \multicolumn{1}{c|}{AP\_M} & AP\_L  \\ \hline
\multicolumn{1}{c|}{FasterRCNN \cite{ren2015faster} }      & \multicolumn{1}{c|}{0.505} & \multicolumn{1}{c|}{0.715} & \multicolumn{1}{c|}{0.594} & \multicolumn{1}{c|}{0.389} & \multicolumn{1}{c|}{0.467} & \multicolumn{1}{c|}{0.539} & \multicolumn{1}{c|}{0.338} & \multicolumn{1}{c|}{0.516} & \multicolumn{1}{c|}{0.390}  & \multicolumn{1}{c|}{0.286}  & \multicolumn{1}{c|}{0.384} & 0.314  \\ \hline
\multicolumn{1}{c|}{CornerNet \cite{law2018cornernet}}         & \multicolumn{1}{c|}{0.410}  & \multicolumn{1}{c|}{0.521} & \multicolumn{1}{c|}{0.439} & \multicolumn{1}{c|}{0.203} & \multicolumn{1}{c|}{0.473} & \multicolumn{1}{c|}{0.390}  & \multicolumn{1}{c|}{0.398} & \multicolumn{1}{c|}{0.517} & \multicolumn{1}{c|}{0.422} & \multicolumn{1}{c|}{0.181}  & \multicolumn{1}{c|}{0.419} & 0.404  \\ \hline
\multicolumn{1}{c|}{CascadeRPN \cite{vu2019cascade} }        & \multicolumn{1}{c|}{0.545} & \multicolumn{1}{c|}{0.715} & \multicolumn{1}{c|}{0.620}  & \multicolumn{1}{c|}{0.385} & \multicolumn{1}{c|}{0.493} & \multicolumn{1}{c|}{0.591} & \multicolumn{1}{c|}{0.412} & \multicolumn{1}{c|}{0.574} & \multicolumn{1}{c|}{0.471} & \multicolumn{1}{c|}{0.316}  & \multicolumn{1}{c|}{0.379} & 0.441  \\ \hline
\multicolumn{1}{c|}{CenterNet \cite{duan2019centernet} }         & \multicolumn{1}{c|}{0.021} & \multicolumn{1}{c|}{0.040}  & \multicolumn{1}{c|}{0.021} & \multicolumn{1}{c|}{0.017} & \multicolumn{1}{c|}{0.018} & \multicolumn{1}{c|}{0.036} & \multicolumn{1}{c|}{0.076} & \multicolumn{1}{c|}{0.161} & \multicolumn{1}{c|}{0.060}  & \multicolumn{1}{c|}{0.048}  & \multicolumn{1}{c|}{0.102} & 0.076  \\ \hline
\multicolumn{1}{c|}{DETR \cite{detr2020}}              & \multicolumn{1}{c|}{0.287} & \multicolumn{1}{c|}{0.506} & \multicolumn{1}{c|}{0.292} & \multicolumn{1}{c|}{0.098} & \multicolumn{1}{c|}{0.201} & \multicolumn{1}{c|}{0.373} & \multicolumn{1}{c|}{0.18}  & \multicolumn{1}{c|}{0.341} & \multicolumn{1}{c|}{0.173} & \multicolumn{1}{c|}{0.053}  & \multicolumn{1}{c|}{0.129} & 0.220   \\ \hline
\multicolumn{1}{c|}{EfficientNet \cite{DBLP:conf/icml/TanL19}}      & \multicolumn{1}{c|}{0.201} & \multicolumn{1}{c|}{0.345} & \multicolumn{1}{c|}{0.225} & \multicolumn{1}{c|}{0.119} & \multicolumn{1}{c|}{0.193} & \multicolumn{1}{c|}{0.218} & \multicolumn{1}{c|}{0.015} & \multicolumn{1}{c|}{0.045} & \multicolumn{1}{c|}{0.004} & \multicolumn{1}{c|}{0.005}  & \multicolumn{1}{c|}{0.016} & 0.014  \\ \hline
\multicolumn{1}{c|}{Deformable-DeTR \cite{zhu2021deformable}} & \multicolumn{1}{c|}{0.550}  & \multicolumn{1}{c|}{0.741} & \multicolumn{1}{c|}{0.645} & \multicolumn{1}{c|}{0.362} & \multicolumn{1}{c|}{0.476} & \multicolumn{1}{c|}{0.612} & \multicolumn{1}{c|}{0.463} & \multicolumn{1}{c|}{0.649} & \multicolumn{1}{c|}{0.538} & \multicolumn{1}{c|}{0.266}  & \multicolumn{1}{c|}{0.42}  & 0.509  \\ \hline
\multicolumn{1}{c|}{YOLOx \cite{DBLP:journals/corr/abs-2107-08430}}             & \multicolumn{1}{c|}{0.332} & \multicolumn{1}{c|}{0.560}  & \multicolumn{1}{c|}{0.356} & \multicolumn{1}{c|}{0.253} & \multicolumn{1}{c|}{0.365} & \multicolumn{1}{c|}{0.312} & \multicolumn{1}{c|}{0.410}  & \multicolumn{1}{c|}{0.595} & \multicolumn{1}{c|}{0.462} & \multicolumn{1}{c|}{0.253}  & \multicolumn{1}{c|}{0.371} & 0.449  \\ \hline
\multicolumn{1}{c|}{YOLOv5s \cite{glenn_jocher_2022_7002879} }           & \multicolumn{1}{c|}{0.388} & \multicolumn{1}{c|}{0.615} & \multicolumn{1}{c|}{0.429} & \multicolumn{1}{c|}{0.301} & \multicolumn{1}{c|}{0.406} & \multicolumn{1}{c|}{0.391} & \multicolumn{1}{c|}{0.510}  & \multicolumn{1}{c|}{0.686} & \multicolumn{1}{c|}{0.600}   & \multicolumn{1}{c|}{0.285}  & \multicolumn{1}{c|}{0.418} & 0.575  \\ \hline
\multicolumn{1}{c|}{DiffusionDet  \cite{chen2023diffusiondet} }      & \multicolumn{1}{c|}{\textbf{0.652}} & \multicolumn{1}{c|}{\textbf{0.792}} & \multicolumn{1}{c|}{\textbf{0.708}} & \multicolumn{1}{c|}{\textbf{0.488}} & \multicolumn{1}{c|}{\textbf{0.580}}  & \multicolumn{1}{c|}{\textbf{0.714}} & \multicolumn{1}{c|}{0.549} & \multicolumn{1}{c|}{0.681} & \multicolumn{1}{c|}{0.599} & \multicolumn{1}{c|}{\textbf{0.405}}  & \multicolumn{1}{c|}{\textbf{0.510}}  & 0.582  \\ \hline
\multicolumn{1}{c|}{YOLOv8 \cite{ultralyticsgithub}}            & \multicolumn{1}{c|}{0.370}  & \multicolumn{1}{c|}{0.578} & \multicolumn{1}{c|}{0.412} & \multicolumn{1}{c|}{0.296} & \multicolumn{1}{c|}{0.390}  & \multicolumn{1}{c|}{0.368} & \multicolumn{1}{c|}{\textbf{0.556}} & \multicolumn{1}{c|}{\textbf{0.692}} & \multicolumn{1}{c|}{\textbf{0.615}} & \multicolumn{1}{c|}{0.344}  & \multicolumn{1}{c|}{0.470}  & \textbf{0.615}  \\ \toprule
\end{tabular}
\label{tab7}
\end{table*}

\begin{table*}[!t]\footnotesize
\centering
\caption{The object detection results of V1-Train [\YellowRect, \BlueRect, \PurpleRect]) and V2-Train [\YellowRect, \BlueRect]) for 11 state-of-the-art detectors on the \datasetname, \emph{w.r.t.}, \textbf{buses}, \textbf{traffic lights}, and \textbf{cyclists}. }
\begin{tabular}{c|cccccc||cccccc}
\multicolumn{13}{c}{\textbf{bus}}                                                                                                                                                                       \\     \toprule
\multicolumn{1}{c}{}                  & \multicolumn{6}{|c|}{V1-Train [\YellowRect, \BlueRect, \PurpleRect], test. [\YellowRect, \BlueRect, \PurpleRect]}                                                                                                                                        & \multicolumn{6}{c}{V2-Train [\YellowRect, \BlueRect], test. [\PurpleRect]}                                                                                                                     \\ \hline
\multicolumn{1}{c|}{Detectors}            & \multicolumn{1}{c|}{AP}   & \multicolumn{1}{c|}{AP50} & \multicolumn{1}{c|}{AP75} & \multicolumn{1}{c|}{AP\_S} & \multicolumn{1}{c|}{AP\_M} & \multicolumn{1}{c|}{AP\_L} & \multicolumn{1}{c|}{AP}   & \multicolumn{1}{c|}{AP50} & \multicolumn{1}{c|}{AP75} & \multicolumn{1}{c|}{AP\_S} & \multicolumn{1}{c|}{AP\_M} & AP\_L \\ \hline
\multicolumn{1}{c|}{FasterRCNN \cite{ren2015faster}}      & \multicolumn{1}{c|}{0.521} & \multicolumn{1}{c|}{0.690}  & \multicolumn{1}{c|}{0.615} & \multicolumn{1}{c|}{0.304} & \multicolumn{1}{c|}{0.431} & \multicolumn{1}{c|}{0.580}  & \multicolumn{1}{c|}{0.312} & \multicolumn{1}{c|}{0.455} & \multicolumn{1}{c|}{0.356} & \multicolumn{1}{c|}{0.263} & \multicolumn{1}{c|}{0.298} & 0.328 \\ \hline
\multicolumn{1}{c|}{CornerNet \cite{law2018cornernet}}         & \multicolumn{1}{c|}{0.380}  & \multicolumn{1}{c|}{0.465} & \multicolumn{1}{c|}{0.408} & \multicolumn{1}{c|}{0.174} & \multicolumn{1}{c|}{0.404} & \multicolumn{1}{c|}{0.376} & \multicolumn{1}{c|}{0.412} & \multicolumn{1}{c|}{0.507} & \multicolumn{1}{c|}{0.443} & \multicolumn{1}{c|}{0.154} & \multicolumn{1}{c|}{0.359} & 0.461 \\ \hline
\multicolumn{1}{c|}{CascadeRPN \cite{vu2019cascade} }        & \multicolumn{1}{c|}{0.522} & \multicolumn{1}{c|}{0.658} & \multicolumn{1}{c|}{0.604} & \multicolumn{1}{c|}{0.263} & \multicolumn{1}{c|}{0.449} & \multicolumn{1}{c|}{0.579} & \multicolumn{1}{c|}{0.395} & \multicolumn{1}{c|}{0.529} & \multicolumn{1}{c|}{0.464} & \multicolumn{1}{c|}{0.214} & \multicolumn{1}{c|}{0.342} & 0.441 \\ \hline
\multicolumn{1}{c|}{CenterNet \cite{duan2019centernet} }         & \multicolumn{1}{c|}{0.003} & \multicolumn{1}{c|}{0.005} & \multicolumn{1}{c|}{0.003} & \multicolumn{1}{c|}{0.001} & \multicolumn{1}{c|}{0.002} & \multicolumn{1}{c|}{0.003} & \multicolumn{1}{c|}{0.027} & \multicolumn{1}{c|}{0.052} & \multicolumn{1}{c|}{0.025} & \multicolumn{1}{c|}{0.028} & \multicolumn{1}{c|}{0.036} & 0.026 \\ \hline
\multicolumn{1}{c|}{DETR \cite{detr2020}}              & \multicolumn{1}{c|}{0.201} & \multicolumn{1}{c|}{0.321} & \multicolumn{1}{c|}{0.219} & \multicolumn{1}{c|}{0.042} & \multicolumn{1}{c|}{0.118} & \multicolumn{1}{c|}{0.258} & \multicolumn{1}{c|}{0.131} & \multicolumn{1}{c|}{0.212} & \multicolumn{1}{c|}{0.141} & \multicolumn{1}{c|}{0.001} & \multicolumn{1}{c|}{0.076} & 0.167 \\ \hline
\multicolumn{1}{c|}{EfficientNet \cite{DBLP:conf/icml/TanL19}}      & \multicolumn{1}{c|}{0.106} & \multicolumn{1}{c|}{0.169} & \multicolumn{1}{c|}{0.123} & \multicolumn{1}{c|}{0.028} & \multicolumn{1}{c|}{0.108} & \multicolumn{1}{c|}{0.109} & \multicolumn{1}{c|}{0.003} & \multicolumn{1}{c|}{0.008} & \multicolumn{1}{c|}{0.001} & \multicolumn{1}{c|}{0.001} & \multicolumn{1}{c|}{0.002} & 0.003 \\ \hline
\multicolumn{1}{c|}{Deformable-DeTR \cite{zhu2021deformable}} & \multicolumn{1}{c|}{0.511} & \multicolumn{1}{c|}{0.670}  & \multicolumn{1}{c|}{0.603} & \multicolumn{1}{c|}{0.266} & \multicolumn{1}{c|}{0.401} & \multicolumn{1}{c|}{0.591} & \multicolumn{1}{c|}{0.484} & \multicolumn{1}{c|}{0.625} & \multicolumn{1}{c|}{0.575} & \multicolumn{1}{c|}{0.282} & \multicolumn{1}{c|}{0.396} & 0.541 \\ \hline
\multicolumn{1}{c|}{YOLOx \cite{DBLP:journals/corr/abs-2107-08430}}             & \multicolumn{1}{c|}{0.595} & \multicolumn{1}{c|}{0.730}  & \multicolumn{1}{c|}{0.678} & \multicolumn{1}{c|}{0.336} & \multicolumn{1}{c|}{0.479} & \multicolumn{1}{c|}{0.670}  & \multicolumn{1}{c|}{0.417} & \multicolumn{1}{c|}{0.556} & \multicolumn{1}{c|}{0.483} & \multicolumn{1}{c|}{0.136} & \multicolumn{1}{c|}{0.33}  & 0.475 \\ \hline
\multicolumn{1}{c|}{YOLOv5s \cite{glenn_jocher_2022_7002879} }           & \multicolumn{1}{c|}{\textbf{0.685}} & \multicolumn{1}{c|}{\textbf{0.794}} & \multicolumn{1}{c|}{\textbf{0.757}} & \multicolumn{1}{c|}{\textbf{0.400}}   & \multicolumn{1}{c|}{\textbf{0.541}} & \multicolumn{1}{c|}{\textbf{0.767}} & \multicolumn{1}{c|}{0.418} & \multicolumn{1}{c|}{0.553} & \multicolumn{1}{c|}{0.503} & \multicolumn{1}{c|}{0.006} & \multicolumn{1}{c|}{0.238} & 0.541 \\ \hline
\multicolumn{1}{c|}{DiffusionDet  \cite{chen2023diffusiondet}}      & \multicolumn{1}{c|}{0.650}  & \multicolumn{1}{c|}{0.759} & \multicolumn{1}{c|}{0.707} & \multicolumn{1}{c|}{0.360}  & \multicolumn{1}{c|}{0.531} & \multicolumn{1}{c|}{0.721} & \multicolumn{1}{c|}{\textbf{0.574}} & \multicolumn{1}{c|}{\textbf{0.674}} & \multicolumn{1}{c|}{\textbf{0.632}}   & \multicolumn{1}{c|}{\textbf{0.315}} & \multicolumn{1}{c|}{\textbf{0.492}} & \textbf{0.631} \\ \hline
\multicolumn{1}{c|}{YOLOv8 \cite{ultralyticsgithub}}            & \multicolumn{1}{c|}{0.668} & \multicolumn{1}{c|}{0.779} & \multicolumn{1}{c|}{0.734} & \multicolumn{1}{c|}{0.371} & \multicolumn{1}{c|}{0.526} & \multicolumn{1}{c|}{0.753} & \multicolumn{1}{c|}{0.533} & \multicolumn{1}{c|}{0.637} & \multicolumn{1}{c|}{0.592} & \multicolumn{1}{c|}{0.123} & \multicolumn{1}{c|}{0.409} & 0.616 \\ 
\toprule
\multicolumn{13}{c}{\textbf{traffic light}}                                                                                                                                                             \\     \toprule
\multicolumn{1}{c}{}                  & \multicolumn{6}{|c|}{V1-Train [\YellowRect, \BlueRect, \PurpleRect], test. [\YellowRect, \BlueRect, \PurpleRect]}                                                                                                                                        & \multicolumn{6}{c}{V2-Train [\YellowRect, \BlueRect], test. [\PurpleRect]}                                                                                                                     \\ \hline
\multicolumn{1}{c|}{Detectors}            & \multicolumn{1}{c|}{AP}   & \multicolumn{1}{c|}{AP50} & \multicolumn{1}{c|}{AP75} & \multicolumn{1}{c|}{AP\_S} & \multicolumn{1}{c|}{AP\_M} & \multicolumn{1}{c|}{AP\_L} & \multicolumn{1}{c|}{AP}   & \multicolumn{1}{c|}{AP50} & \multicolumn{1}{c|}{AP75} & \multicolumn{1}{c|}{AP\_S} & \multicolumn{1}{c|}{AP\_M} & AP\_L \\ \hline
\multicolumn{1}{c|}{FasterRCNN \cite{ren2015faster}}      & \multicolumn{1}{c|}{0.487} & \multicolumn{1}{c|}{0.689} & \multicolumn{1}{c|}{0.583} & \multicolumn{1}{c|}{0.434} & \multicolumn{1}{c|}{0.515} & \multicolumn{1}{c|}{0.208} & \multicolumn{1}{c|}{0.371} & \multicolumn{1}{c|}{0.528} & \multicolumn{1}{c|}{0.451} & \multicolumn{1}{c|}{0.325} & \multicolumn{1}{c|}{0.402} & 0.039 \\ \hline
\multicolumn{1}{c|}{CornerNet \cite{law2018cornernet}}         & \multicolumn{1}{c|}{0.412} & \multicolumn{1}{c|}{0.543} & \multicolumn{1}{c|}{0.482} & \multicolumn{1}{c|}{0.306} & \multicolumn{1}{c|}{0.506} & \multicolumn{1}{c|}{0.024} & \multicolumn{1}{c|}{0.248} & \multicolumn{1}{c|}{0.317} & \multicolumn{1}{c|}{0.280}  & \multicolumn{1}{c|}{0.275} & \multicolumn{1}{c|}{0.286} & 0.018 \\ \hline
\multicolumn{1}{c|}{CascadeRPN \cite{vu2019cascade} }        & \multicolumn{1}{c|}{0.495} & \multicolumn{1}{c|}{0.675} & \multicolumn{1}{c|}{0.585} & \multicolumn{1}{c|}{0.417} & \multicolumn{1}{c|}{0.532} & \multicolumn{1}{c|}{0.226} & \multicolumn{1}{c|}{0.409} & \multicolumn{1}{c|}{0.531} & \multicolumn{1}{c|}{0.480}  & \multicolumn{1}{c|}{0.368} & \multicolumn{1}{c|}{0.437} & 0.103 \\ \hline
\multicolumn{1}{c|}{CenterNet \cite{duan2019centernet} }         & \multicolumn{1}{c|}{0.061} & \multicolumn{1}{c|}{0.127} & \multicolumn{1}{c|}{0.048} & \multicolumn{1}{c|}{0.040}  & \multicolumn{1}{c|}{0.085} & \multicolumn{1}{c|}{0.000}     & \multicolumn{1}{c|}{0.076} & \multicolumn{1}{c|}{0.167} & \multicolumn{1}{c|}{0.057} & \multicolumn{1}{c|}{0.070}  & \multicolumn{1}{c|}{0.094} & 0.000     \\ \hline
\multicolumn{1}{c|}{DETR \cite{detr2020}}              & \multicolumn{1}{c|}{0.132} & \multicolumn{1}{c|}{0.359} & \multicolumn{1}{c|}{0.069} & \multicolumn{1}{c|}{0.062} & \multicolumn{1}{c|}{0.163} & \multicolumn{1}{c|}{0.068} & \multicolumn{1}{c|}{0.079} & \multicolumn{1}{c|}{0.243} & \multicolumn{1}{c|}{0.024} & \multicolumn{1}{c|}{0.048} & \multicolumn{1}{c|}{0.095} & 0.046 \\ \hline
\multicolumn{1}{c|}{EfficientNet \cite{DBLP:conf/icml/TanL19}}      & \multicolumn{1}{c|}{0.164} & \multicolumn{1}{c|}{0.260}  & \multicolumn{1}{c|}{0.169} & \multicolumn{1}{c|}{0.090}  & \multicolumn{1}{c|}{0.207} & \multicolumn{1}{c|}{0.011} & \multicolumn{1}{c|}{0.024} & \multicolumn{1}{c|}{0.089} & \multicolumn{1}{c|}{0.000}     & \multicolumn{1}{c|}{0.006} & \multicolumn{1}{c|}{0.033} & 0.000     \\ \hline
\multicolumn{1}{c|}{Deformable-DeTR \cite{zhu2021deformable}} & \multicolumn{1}{c|}{0.394} & \multicolumn{1}{c|}{0.669} & \multicolumn{1}{c|}{0.450}  & \multicolumn{1}{c|}{0.345} & \multicolumn{1}{c|}{0.420}  & \multicolumn{1}{c|}{0.304} & \multicolumn{1}{c|}{0.320}  & \multicolumn{1}{c|}{0.585} & \multicolumn{1}{c|}{0.323} & \multicolumn{1}{c|}{0.264} & \multicolumn{1}{c|}{0.35}  & 0.155 \\ \hline
\multicolumn{1}{c|}{YOLOx \cite{DBLP:journals/corr/abs-2107-08430}}             & \multicolumn{1}{c|}{0.480}  & \multicolumn{1}{c|}{0.667} & \multicolumn{1}{c|}{0.570}  & \multicolumn{1}{c|}{0.384} & \multicolumn{1}{c|}{0.546} & \multicolumn{1}{c|}{0.328} & \multicolumn{1}{c|}{0.310}  & \multicolumn{1}{c|}{0.458} & \multicolumn{1}{c|}{0.359} & \multicolumn{1}{c|}{0.248} & \multicolumn{1}{c|}{0.346} & 0.151 \\ \hline
\multicolumn{1}{c|}{YOLOv5s \cite{glenn_jocher_2022_7002879} }           & \multicolumn{1}{c|}{\textbf{0.542}} & \multicolumn{1}{c|}{\textbf{0.743}} & \multicolumn{1}{c|}{\textbf{0.653}} & \multicolumn{1}{c|}{0.428} & \multicolumn{1}{c|}{\textbf{0.590}}  & \multicolumn{1}{c|}{0.423} & \multicolumn{1}{c|}{0.356} & \multicolumn{1}{c|}{0.548} & \multicolumn{1}{c|}{0.420}  & \multicolumn{1}{c|}{0.297} & \multicolumn{1}{c|}{0.391} & 0.207 \\ \hline
\multicolumn{1}{c|}{DiffusionDet  \cite{chen2023diffusiondet}}      & \multicolumn{1}{c|}{0.522} & \multicolumn{1}{c|}{0.703} & \multicolumn{1}{c|}{0.605} & \multicolumn{1}{c|}{\textbf{0.440}}  & \multicolumn{1}{c|}{0.570}  & \multicolumn{1}{c|}{0.344} & \multicolumn{1}{c|}{\textbf{0.511}} & \multicolumn{1}{c|}{\textbf{0.680}}  & \multicolumn{1}{c|}{\textbf{0.589}} & \multicolumn{1}{c|}{\textbf{0.441}} & \multicolumn{1}{c|}{\textbf{0.559}} & \textbf{0.248} \\ \hline
\multicolumn{1}{c|}{YOLOv8 \cite{ultralyticsgithub}}            & \multicolumn{1}{c|}{0.526} & \multicolumn{1}{c|}{0.703} & \multicolumn{1}{c|}{0.622} & \multicolumn{1}{c|}{0.413} & \multicolumn{1}{c|}{0.575} & \multicolumn{1}{c|}{\textbf{0.429}} & \multicolumn{1}{c|}{0.417} & \multicolumn{1}{c|}{0.570}  & \multicolumn{1}{c|}{0.492} & \multicolumn{1}{c|}{0.317} & \multicolumn{1}{c|}{0.465} & 0.247 \\ 
\toprule
\multicolumn{13}{c}{\textbf{cyclist}}                                                                                                                                                                   \\     \toprule
\multicolumn{1}{c}{}                  & \multicolumn{6}{|c|}{V1-Train [\YellowRect, \BlueRect, \PurpleRect], test. [\YellowRect, \BlueRect, \PurpleRect]}                                                                                                                                        & \multicolumn{6}{c}{V2-Train [\YellowRect, \BlueRect], test. [\PurpleRect]}                                                                                                                     \\ \hline
\multicolumn{1}{c|}{Detectors}            & \multicolumn{1}{c|}{AP}   & \multicolumn{1}{c|}{AP50} & \multicolumn{1}{c|}{AP75} & \multicolumn{1}{c|}{AP\_S} & \multicolumn{1}{c|}{AP\_M} & \multicolumn{1}{c|}{AP\_L} & \multicolumn{1}{c|}{AP}   & \multicolumn{1}{c|}{AP50} & \multicolumn{1}{c|}{AP75} & \multicolumn{1}{c|}{AP\_S} & \multicolumn{1}{c|}{AP\_M} & AP\_L \\ \hline
\multicolumn{1}{c|}{FasterRCNN \cite{ren2015faster}}      & \multicolumn{1}{c|}{0.218} & \multicolumn{1}{c|}{0.391} & \multicolumn{1}{c|}{0.246} & \multicolumn{1}{c|}{0.015} & \multicolumn{1}{c|}{0.248} & \multicolumn{1}{c|}{0.196} & \multicolumn{1}{c|}{0.122} & \multicolumn{1}{c|}{0.255} & \multicolumn{1}{c|}{0.105} & \multicolumn{1}{c|}{0.086} & \multicolumn{1}{c|}{0.144} & 0.072 \\ \hline
\multicolumn{1}{c|}{CornerNet \cite{law2018cornernet}}         & \multicolumn{1}{c|}{0.179} & \multicolumn{1}{c|}{0.297} & \multicolumn{1}{c|}{0.184} & \multicolumn{1}{c|}{0.020}  & \multicolumn{1}{c|}{0.196} & \multicolumn{1}{c|}{0.191} & \multicolumn{1}{c|}{0.227} & \multicolumn{1}{c|}{0.370}  & \multicolumn{1}{c|}{0.242} & \multicolumn{1}{c|}{0.034} & \multicolumn{1}{c|}{0.257} & 0.194 \\ \hline
\multicolumn{1}{c|}{CascadeRPN \cite{vu2019cascade} }        & \multicolumn{1}{c|}{0.255} & \multicolumn{1}{c|}{0.446} & \multicolumn{1}{c|}{0.276} & \multicolumn{1}{c|}{0.017} & \multicolumn{1}{c|}{0.253} & \multicolumn{1}{c|}{0.318} & \multicolumn{1}{c|}{0.150}  & \multicolumn{1}{c|}{0.275} & \multicolumn{1}{c|}{0.166} & \multicolumn{1}{c|}{0.066} & \multicolumn{1}{c|}{0.159} & 0.171 \\ \hline
\multicolumn{1}{c|}{CenterNet \cite{duan2019centernet} }         & \multicolumn{1}{c|}{0.000}     & \multicolumn{1}{c|}{0.000}     & \multicolumn{1}{c|}{0.000}     & \multicolumn{1}{c|}{0.000}     & \multicolumn{1}{c|}{0.000}     & \multicolumn{1}{c|}{0.000}     & \multicolumn{1}{c|}{0.000}     & \multicolumn{1}{c|}{0.000}     & \multicolumn{1}{c|}{0.000}     & \multicolumn{1}{c|}{0.001} & \multicolumn{1}{c|}{0.000}     & 0.000     \\ \hline
\multicolumn{1}{c|}{DETR \cite{detr2020}}              & \multicolumn{1}{c|}{0.035} & \multicolumn{1}{c|}{0.106} & \multicolumn{1}{c|}{0.011} & \multicolumn{1}{c|}{0.012} & \multicolumn{1}{c|}{0.033} & \multicolumn{1}{c|}{0.051} & \multicolumn{1}{c|}{0.003} & \multicolumn{1}{c|}{0.012} & \multicolumn{1}{c|}{0.002} & \multicolumn{1}{c|}{0.000}     & \multicolumn{1}{c|}{0.003} & 0.007 \\ \hline
\multicolumn{1}{c|}{EfficientNet \cite{DBLP:conf/icml/TanL19}}      & \multicolumn{1}{c|}{0.016} & \multicolumn{1}{c|}{0.039} & \multicolumn{1}{c|}{0.010}  & \multicolumn{1}{c|}{0.002} & \multicolumn{1}{c|}{0.018} & \multicolumn{1}{c|}{0.019} & \multicolumn{1}{c|}{0.000}     & \multicolumn{1}{c|}{0.000}     & \multicolumn{1}{c|}{0.000}     & \multicolumn{1}{c|}{0.000}     & \multicolumn{1}{c|}{0.000}     & 0.000     \\ \hline
\multicolumn{1}{c|}{Deformable-DeTR \cite{zhu2021deformable}} & \multicolumn{1}{c|}{0.231} & \multicolumn{1}{c|}{0.446} & \multicolumn{1}{c|}{0.221} & \multicolumn{1}{c|}{0.038} & \multicolumn{1}{c|}{0.244} & \multicolumn{1}{c|}{0.255} & \multicolumn{1}{c|}{0.166} & \multicolumn{1}{c|}{0.313} & \multicolumn{1}{c|}{0.158} & \multicolumn{1}{c|}{0.088} & \multicolumn{1}{c|}{0.177} & 0.161 \\ \hline
\multicolumn{1}{c|}{YOLOx \cite{DBLP:journals/corr/abs-2107-08430}}             & \multicolumn{1}{c|}{0.235} & \multicolumn{1}{c|}{0.439} & \multicolumn{1}{c|}{0.223} & \multicolumn{1}{c|}{0.048} & \multicolumn{1}{c|}{0.228} & \multicolumn{1}{c|}{0.309} & \multicolumn{1}{c|}{0.137} & \multicolumn{1}{c|}{0.271} & \multicolumn{1}{c|}{0.121} & \multicolumn{1}{c|}{0.081} & \multicolumn{1}{c|}{0.174} & 0.108 \\ \hline
\multicolumn{1}{c|}{YOLOv5s \cite{glenn_jocher_2022_7002879} }           & \multicolumn{1}{c|}{\textbf{0.368}} & \multicolumn{1}{c|}{\textbf{0.601}} & \multicolumn{1}{c|}{\textbf{0.423}} & \multicolumn{1}{c|}{\textbf{0.058}} & \multicolumn{1}{c|}{0.362} & \multicolumn{1}{c|}{\textbf{0.477}} & \multicolumn{1}{c|}{0.005} & \multicolumn{1}{c|}{0.012} & \multicolumn{1}{c|}{0.004} & \multicolumn{1}{c|}{0.000}     & \multicolumn{1}{c|}{0.006} & 0.004 \\ \hline
\multicolumn{1}{c|}{DiffusionDet  \cite{chen2023diffusiondet}}      & \multicolumn{1}{c|}{0.360}  & \multicolumn{1}{c|}{0.577} & \multicolumn{1}{c|}{0.391} & \multicolumn{1}{c|}{0.036} & \multicolumn{1}{c|}{\textbf{0.373}} & \multicolumn{1}{c|}{0.418} & \multicolumn{1}{c|}{\textbf{0.302}} & \multicolumn{1}{c|}{\textbf{0.474}} & \multicolumn{1}{c|}{\textbf{0.319}} & \multicolumn{1}{c|}{\textbf{0.095}} & \multicolumn{1}{c|}{\textbf{0.318}} & \textbf{0.291} \\ \hline
\multicolumn{1}{c|}{YOLOv8 \cite{ultralyticsgithub}}            & \multicolumn{1}{c|}{0.314} & \multicolumn{1}{c|}{0.508} & \multicolumn{1}{c|}{0.351} & \multicolumn{1}{c|}{0.057} & \multicolumn{1}{c|}{0.302} & \multicolumn{1}{c|}{0.416} & \multicolumn{1}{c|}{0.198} & \multicolumn{1}{c|}{0.332} & \multicolumn{1}{c|}{0.211} & \multicolumn{1}{c|}{0.070}  & \multicolumn{1}{c|}{0.236} & 0.221 \\ \toprule
\end{tabular}
\label{tab8}
\end{table*}
 \begin{figure*}[htpb]
  \centering
\includegraphics[width=\linewidth]{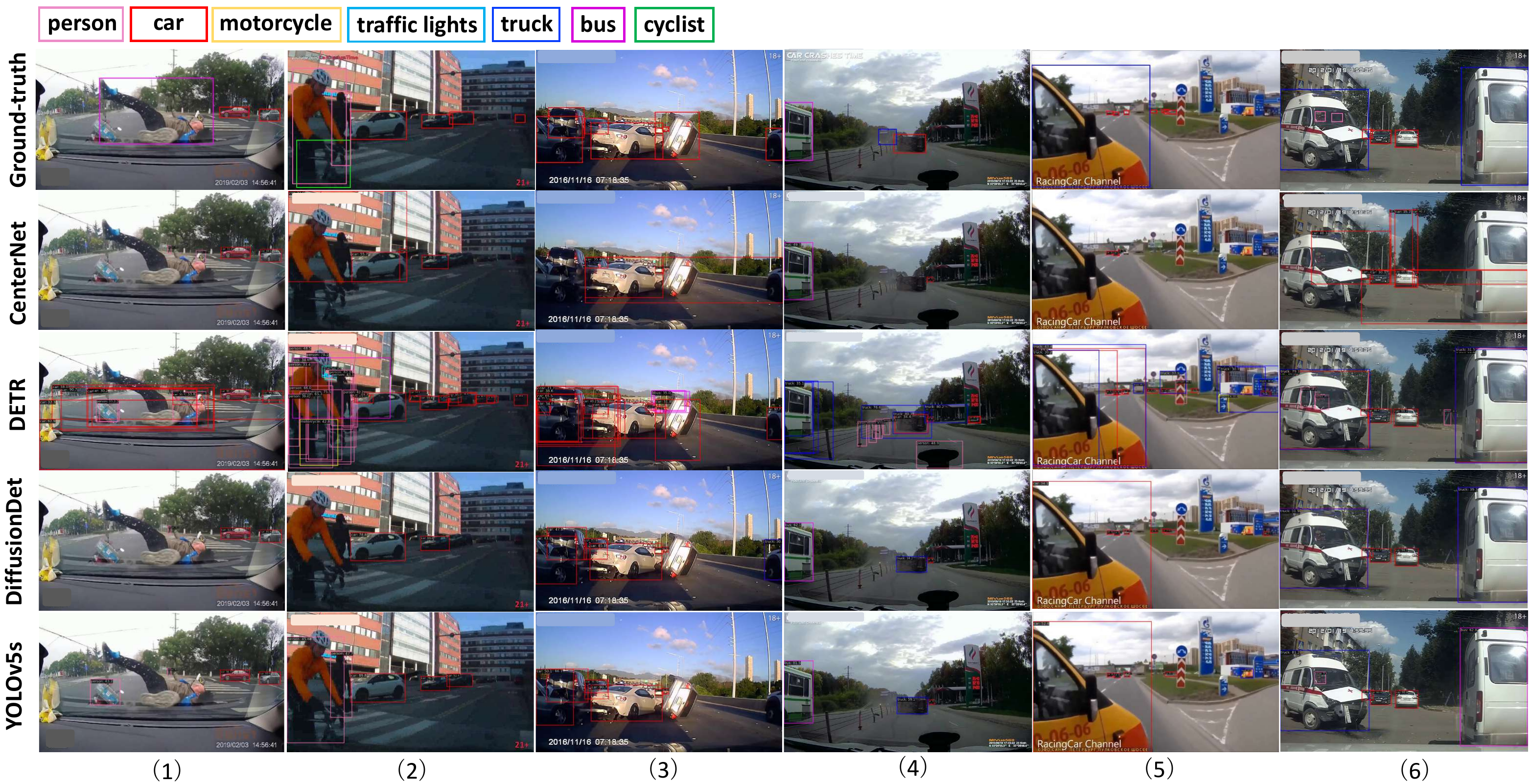}
   \caption{\small{The object detection snapshots in accident frames by CenterNet \cite{law2018cornernet}, DETR \cite{detr2020}, DiffusionDet  \cite{chen2023diffusiondet}, and YOLOv5s\cite{glenn_jocher_2022_7002879}. We can see that all detectors fail to detect the cyclist (column (2)) and the pedestrian with distorted posture (column (1)). DETR is more active for covering all possible objects while many false detections are generated.}}
   \label{fig10}
   \vspace{-0.5em}
\end{figure*}

\section{The Architecture of 3D-CAB}
To be clear for re-reproduction, we detail the workflow of 3D-CAB in OAVD, as shown in Fig. \ref{fig9}. $B$ denotes the batch size, and the maximum text prompt length $L$ is set to 77. In each layer of 3D-CAB, $c$, $h$, and $w$ represent the channels, height, and width of the input video feature $\textbf{z}_{v_{in}}$, and $C$, $H$, and $W$ represent the channels, height, and width of the video clip representation after ResNet Block encoding. Notably, the channels, height, and width in each step of 3D-CAB change for a dimension adaptation. Furthermore, we inject different attention modules, \ie, \textbf{SA}, \textbf{CA}, and \textbf{TA}, into Low-rank Adaptation (LoRA) trainer \footnote{\url{https://github.com/cloneofsimo/lora}} for fast fine-tuning on LDM \cite{rombach2022high}.
 
\section{OD Analysis for Different Kinds of Objects}
For an adequate benchmark, we offer a more detailed Object Detection (OD) analysis for distinct object types.  Likewise, our evaluation utilizes the Average Precision (AP) metrics. In this context, we consider the original AP (average precision with IoU thresholds ranging from 0.5 to 0.95), AP50 (with an IoU threshold of 0.5), and AP75 (with an IoU threshold of 0.75) for our assessment. Additionally, due to the varying scales of the objects involved in collisions during accident scenarios, we have evaluated the proficiency of the model in detecting objects of small ($<32*32$), medium ($>32*32$ \& $<96*96$), and large ($>96*96$) scales, as measured by AP\_S, AP\_M, and AP\_L. 

We present the fine-grained quantitative object analysis for 11 state-of-the-art detectors in Tab. \ref{tab7} and Tab. \ref{tab8}. According to the results, we can see that the accuracy of both detectors, YOLOv5s and DiffusionDet are the best in almost all object categories. YOLOv5s is better than DiffusionDet with V1-Train [\YellowRect, \BlueRect, \PurpleRect] for testing [\YellowRect, \BlueRect, \PurpleRect], while DiffusionDet benefits from excellent generalization (V2-Train [\YellowRect, \BlueRect], test.[\PurpleRect]), which allows DiffusionDet to detect important objects in accident scenarios even if these objects are not present in the training data. 

 \noindent \textbf{Sensitivity to Different Kinds of Objects:}
 According to the results of Tab. \ref{tab7} and Tab. \ref{tab8}, all object detectors perform best when detecting cars as they are the most commonly occurring object in \datasetname. YOLOv5s obtains 0.936 of AP50 in the V1-Train mode, and DiffusionDet generates 0.908 of AP50 under the V2-Train mode. For cars, pedestrians, trucks, buses, and traffic lights, the \textbf{AP} values of the best detector are larger than 0.5. Yet, motorcycles and cyclists are hard to be detected especially under the V2-Train mode, where all kinds of AP values are less than 0.5. Here, compared with DiffusionDet, YOLOv5s is with failure on motorcycles and cyclists in the V2-Train mode.

 \noindent \textbf{Adaptability to Small Objects:}
 Small object detection is a difficult problem because there are not enough details to obtain a strong feature representation. As for accident scenarios, this problem may be aggravated because of the unusual property. Therefore, we can observe that most detectors generate the lowest AP\_S values within their AP value set. For motorcycles, traffic lights, and pedestrians, too large objects commonly are unusual and AP\_L values are the smallest in V2-Train mode. Contrarily, for these kinds of objects, AP\_L values in V1-Train mode are not the smallest, which indicates that the large size of objects in the accident window frequently appears due to the severe scale change, \emph{e.g.,} the ego-car involved cases in Fig. \ref{fig10} (1)-(2) and (5)-(6).

  \noindent \textbf{Scalability to Corner Objects:}
The objects in the road accident window are the typical corner cases in object detection. Fig. \ref{fig10} demonstrates some examples of the detection results of CenterNet, DETR, DiffusionDet, and YOLOv5s. It is clear that these corner cases are hard to address because of the dramatic scale change (Fig. \ref{fig10}(1)-(2) and (5)-(6)) and severe pose distortion (Fig. \ref{fig10}(1) and (3)-(4)). Many objects are wrongly detected, such as the wrong detections of \emph{car}$\rightarrow$\emph{truck}, \emph{bus}$\rightarrow$\emph{truck}. DETR is more active in covering all possible objects while generates many false detections.

In summary, due to the corner cases, object detection in ego-view accident videos still has many unresolved issues.

  \begin{figure*}[htpb]
  \centering
\includegraphics[width=0.85\linewidth]{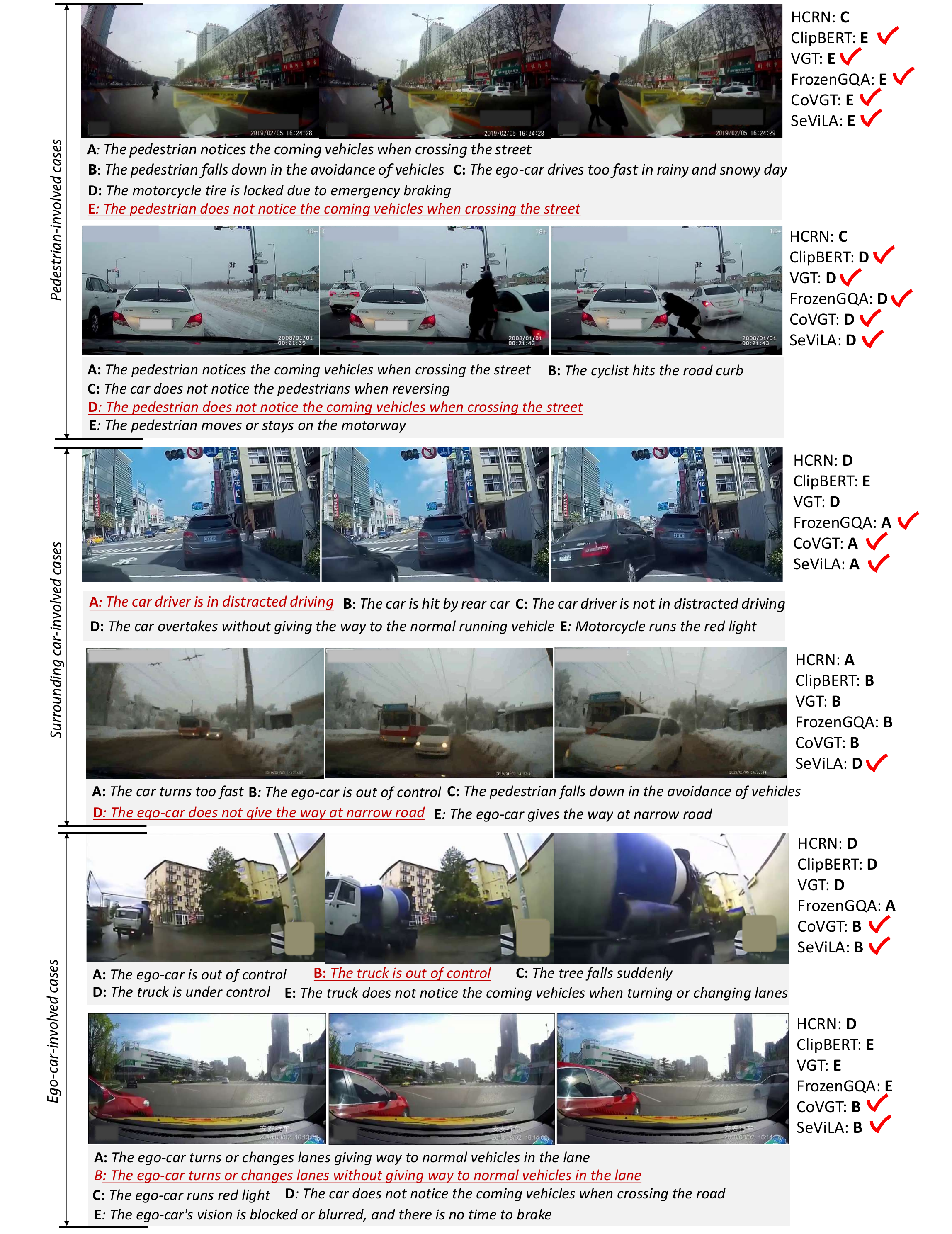}
   \caption{\small{The \textbf{case visualization of Accident reason Answering (ArA)} by 8 state-of-the-art Video Question Answering (VQA) methods.}}
   \label{fig11}
\end{figure*}

\section{ArA Case Analysis, \emph{w.r.t.}, Different Objects}
Continuing the aforementioned analysis of the ArA task in the main body, we show some cases with respect to different objects in Fig. \ref{fig11} from the results of the state-of-the-art methods. We can see that because many pedestrian-involved accidents may be caused by distracted walking or aggressive movement, such as sudden crossing, besides HCRN~\cite{le2020hierarchical}, all the methods can provide an accurate accident reason for the shown cases. For the surrounding car-involved cases, the irregular behaviors of cars are the common reason for the accidents, which implies a traffic rule reasoning problem. Therefore, the methods with better commonsense knowledge learning, such as SeViLA~\cite{yu2023self} (the only method for the accurate ArA for the $4th$ case), have advantages. As for the ego-car involved cases, the severe scale change advocates the object-centric methods with better region context learning.

   \begin{figure*}[htpb]
  \centering
\includegraphics[width=0.95\linewidth]{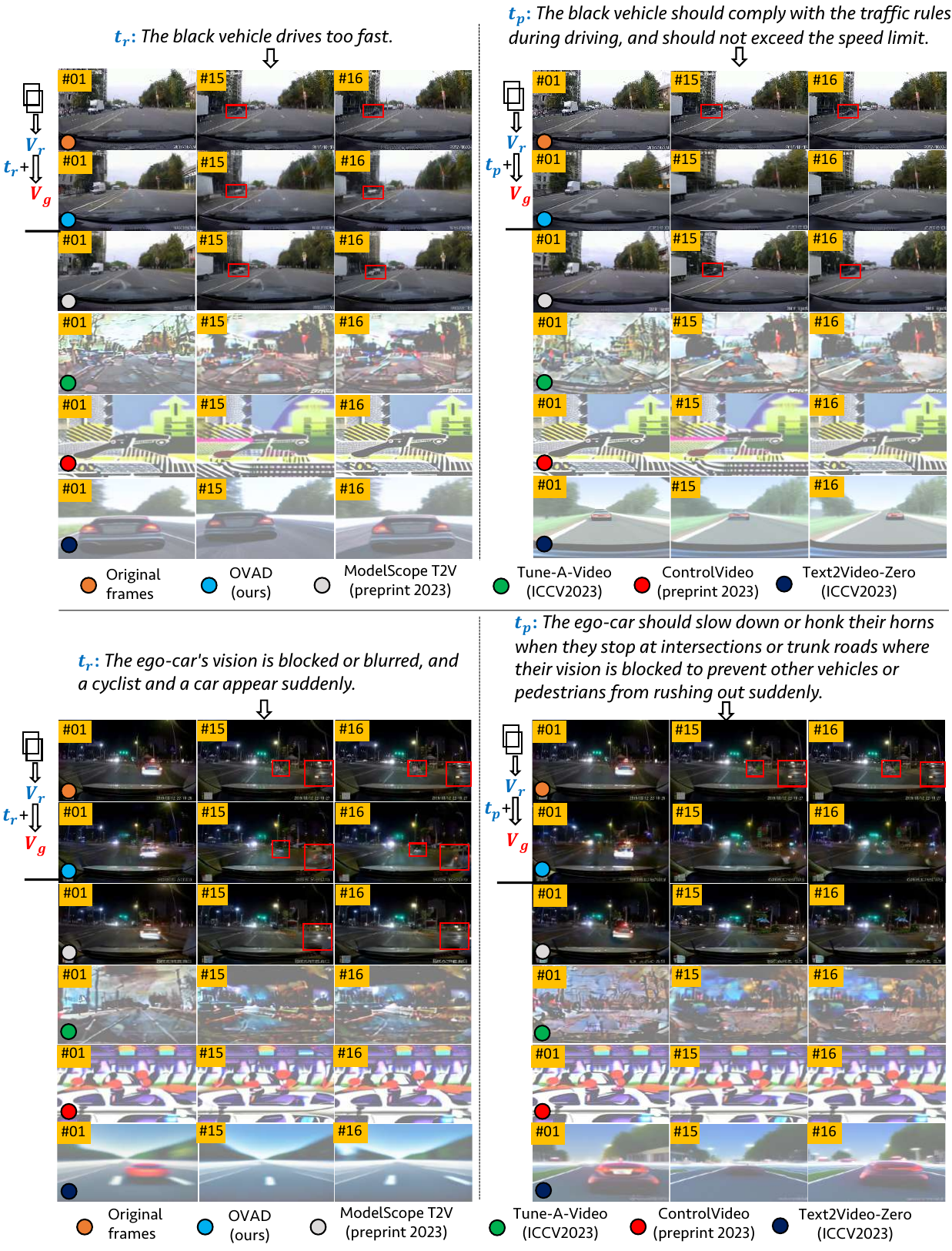}
   \caption{\small{The visualization of generated frames by our OAVD, ModelScope T2V, Tune-A-Video \cite{wu2023tune}, ControlVideo \cite{zhang2023controlvideo}, and Text2Video-Zero.}}
   \label{fig12}
   \vspace{-0.5em}
\end{figure*}

    \begin{figure*}[htpb]
  \centering
\includegraphics[width=0.95\linewidth]{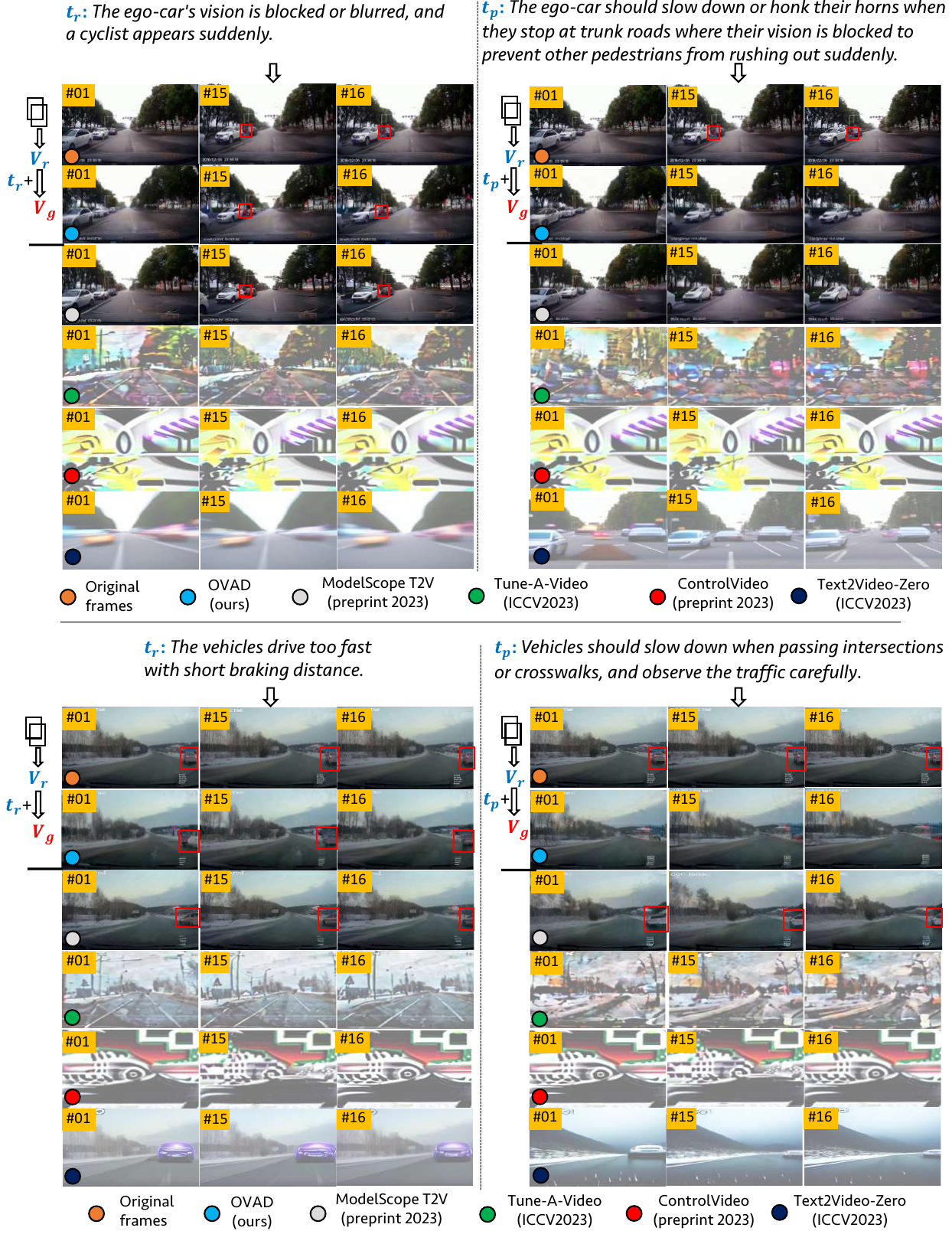}
   \caption{\small{The visualization of video diffusion by our OAVD, ModelScope T2V, Tune-A-Video \cite{wu2023tune}, ControlVideo \cite{zhang2023controlvideo}, and Text2Video-Zero.}}
   \label{fig13}
   \vspace{-0.5em}
\end{figure*}
\section{More Evaluations of OAVD}
More evaluations are provided here for a sufficient understanding of our Object-centric Accident Video Diffusion (OAVD). We provide more example analysis to check the abductive ability by our OAVD with a comparison to other state-of-the-art video diffusion methods. Notably, we further include ModelScope T2V (preprint)\footnote{\url{https://modelscope.cn/models/damo/text-to-video-synthesis/summary}} and Text2Video-Zero (published in ICCV2023)\footnote{\url{https://github.com/Picsart-AI-Research/Text2Video-Zero}} in the evaluation.
ModelScope T2V is re-trained by a same number of samples with our OAVD (\ie, 6000 Co-CPs), and Text2Video-Zero is another training-free video diffusion method.

  \noindent \textbf{More Visualizations of OAVD Against SOTAs:}
Fig. \ref{fig12} and Fig. \ref{fig13} present the qualitative comparisons of different video diffusion models. The inference flow is $(Bboxes\rightarrow$\textcolor{highblue}{$V_{r}$})$+$\textcolor{highblue}{$t_{r}/t_{p}$}$\rightarrow$\textcolor{red}{$V_{g}$}, \emph{i.e.}, that we input the detected bounding boxes $Bboxes$, the video clip in near-accident window \textcolor{highblue}{$V_{r}$} \BlueRect~, and the accident reason or prevention advice description \textcolor{highblue}{$t_{r}/t_{p}$}. From the demonstrated snapshots, we can see that, our OAVD similarly shows an ``\emph{in advance}" phenomenon for the accident reason prompt and eliminates the crashing object when inputting the prevention advice description. ModelScope T2V also generates promising video frames with clear details, even with the ability to eliminate the objects to be involved in accidents after inputting the prevention advice description, as shown by the second example in Fig. \ref{fig12} and the first case in Fig. \ref{fig13}. Yet, it is not stable verified by Fig. \ref{fig12} (the $1st$ example) and Fig. \ref{fig13} (the $2nd$ sample). As for other methods, including the training-free ones, the style and the content of the generated video frames are not relevant to the given text prompts.
   \begin{figure*}[htpb]
  \centering
\includegraphics[width=0.9\linewidth]{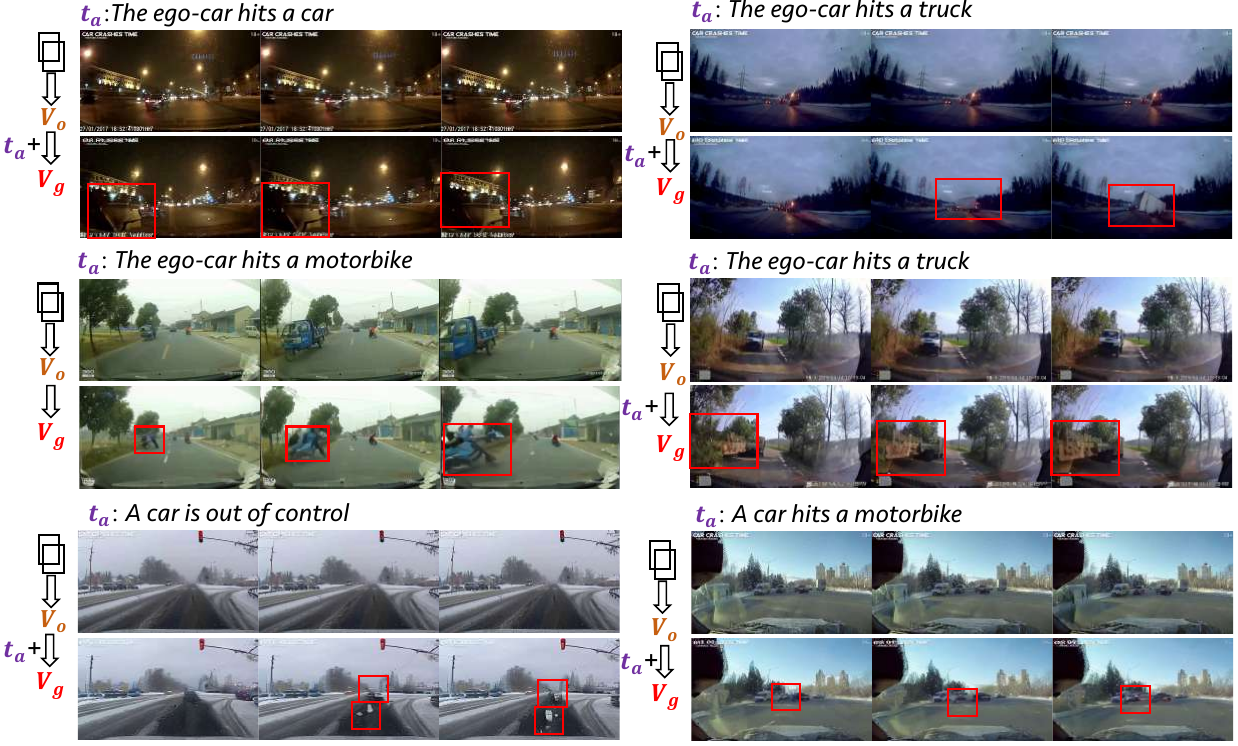}
   \caption{\small{The visualization of \textbf{accident video generation} of OAVD with the inference path of $(Bboxes\rightarrow$\textcolor{earthyellow}{$V_{o}$})$+$\textcolor{magenta}{$t_{a}$}$\rightarrow$\textcolor{red}{$V_{g}$}.}}
   \label{fig14}
   \vspace{-0.5em}
\end{figure*}

    \begin{figure*}[htpb]
  \centering
\includegraphics[width=0.9\linewidth]{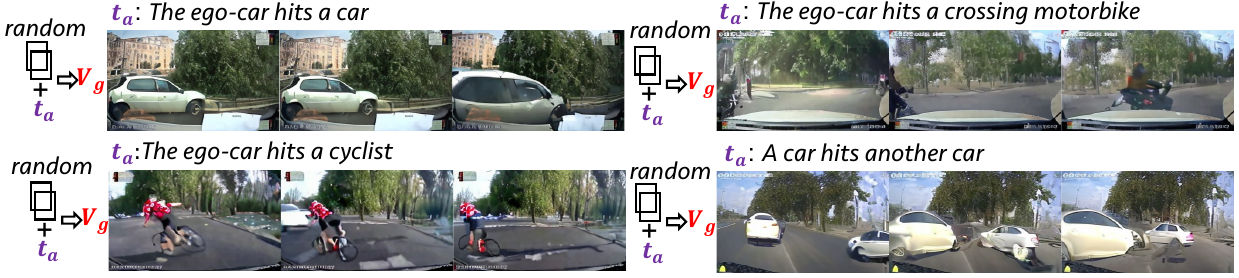}
   \caption{\small{The visualization of \textbf{video-free accident video generation} of OAVD with the inference path of $Bboxes$$+$\textcolor{magenta}{$t_{a}$}$\rightarrow$\textcolor{red}{$V_{g}$}.}}
   \label{fig15}
   \vspace{-0.5em}
\end{figure*}

  \noindent \textbf{More Analysis on the Impact of Bboxes:}
To be clear about the impact of bounding boxes (\textbf{Bboxes}) for our OAVD model, we re-train the OAVD without the input of Bboxes driven by the Sequential CLIP (S-CLIP) and Abductive CLIP (A-CLIP) models. The video-level Fréchet Video Distance (\textbf{FVD}) \cite{unterthiner2018towards} is adopted here. The results in Tab. \ref{tabbb} show that the bounding boxes are useful for enhancing the video quality, and lower FVD values are generated. Based on the evaluation, object-centric video diffusion is promising for generating detailed frame content. 
\begin{table}[htpb]\scriptsize
  \centering
  \caption{\small{FVD value comparison of our OAVD with or without the input of bounding boxes. *: with the input of bounding boxes.}}
      \renewcommand{\arraystretch}{1.2}
 \setlength{\tabcolsep}{0.1mm}{
 \begin{tabular}{c|c|c|c|c}
     \toprule
Method  &OAVD (S-CLIP)$^{*}$&OAVD (S-CLIP)&OAVD (A-CLIP)$^{*}$&OAVD (A-CLIP)\\
\hline
FVD $\downarrow$    & \textbf{5372.3} &5384.6&\textbf{5238.1}&5358.8\\
  \toprule
\end{tabular}}
\vspace{-1em}
\label{tabbb}
\end{table}
 
  \noindent \textbf{Visualizations of Accident Video Generation:}
Besides the abductive check for our video diffusion model OAVD, we also show its ability for flexible accident video generation. To be clear, the inference stage here takes the video clip in normal video segment \textcolor{earthyellow}{$V_o$} \YellowRect~and the accident category description \textcolor{magenta}{$t_a$}. This configuration verifies the reality-changing ability from normal situations to accidents. Fig. \ref{fig14} shows some examples of accident video generation. We can curiously find that our OAVD can create the object to be involved in accidents with a clear pose or appearance. This ability may address the few-shot sample issue of accident videos for future task use.

In addition, we also check the video-free accident video generation by only inputting the bounding boxes to our OAVD. Here, the four \footnote{Other values can also be set.} bounding boxes are randomly set for each example. From the results in Fig. \ref{fig15}, the guidance of the accident category description is clearly verified and the generated accident videos are more realistic without the restriction of original video frames. From these visualizations, OAVD can flexibly augment the video sample scale of ego-view accidents for safe driving.

\end{document}